\documentclass[10pt]{article} 

\usepackage[preprint]{rlj} 

\usepackage{amssymb}            
\usepackage{mathtools}          
\usepackage{mathrsfs}           
\usepackage{graphicx}           
\usepackage{subcaption}         
\usepackage[space]{grffile}     
\usepackage{url}                
\usepackage{lipsum}             
\usepackage{placeins}

\usepackage{booktabs} 
\usepackage{listings}
\usepackage[capitalize,noabbrev]{cleveref}
\usepackage{algorithm}
\usepackage{algorithmic}

\title{Calibrated Partial Resets: Preventing Policy Collapse in Continual Reinforcement Learning}

\setrunningtitle{Calibrated Partial Resets: Preventing Policy Collapse in Continual Reinforcement Learning}

\author{Luc McCutcheon\textsuperscript{1}, Evangelos Chatzaroulas\textsuperscript{2}, Saber Fallah\textsuperscript{1}}

\emails{\{lm01065,e.chatzaroulas,s.fallah\}@surrey.ac.uk}

\affiliations{
$^{1}$\textbf{Department of Mechanical Engineering, University of Surrey, United Kingdom}\\
$^{2}$\textbf{Department of Computing Science, University of Surrey, United Kingdom}\\
}

\contribution{
	 We identify binary neuron resets as a source of instability in long-horizon continual RL and propose CPR, a utility-scaled partial reset optimizer
    }
    {
    Prior work demonstrated targeted \emph{binary} resets as a method of plasticity introduction \citep{cbp,redo,grama}
    }

\contribution{
	We show that CPR avoids policy collapse over 400M steps in SlipperyAnt and improves performance on SlipperyHumanoid, Continual MetaWorld, Continual MinAtar, and time-delayed Ant
    }
    {
The 400M-step friction-shift setting exposes long-horizon instabilities in binary reset methods that are less visible in shorter evaluations.
    }

\contribution{
	We ablate reset magnitude and transformation shape, showing that partial, utility-weighted resets control the plasticity–stability tradeoff.
    }
    {
    None
    }

\keywords{Continual Learning, CRL, stability-plasticity} 

\summary{
	Neural networks are hindered by accumulating dormant neurons and loss of expressivity throughout training, particularly in non-stationary data settings, such as continual supervised and reinforcement learning. Recently, neuron resets have been used to maintain gradient flow and restore plasticity. However, full unit reinitialization often sacrifices peak performance and can destabilize training, leading to policy collapse.
	To preserve plasticity without destabilizing training, we propose Calibrated Partial Resets (CPR), an optimizer that periodically pulls low-utility neurons toward their initialization, with pull strength scaled by each neuron's utility. Unlike binary reset methods, \emph{partial} resets avoid brittleness; unlike uniform decay, \emph{calibrated} utility-scaling concentrates adjustment on the units that need it most.
	Among compared methods, only CPR avoids policy collapse over 400M training steps in SlipperyAnt, and it outperforms prior decay and reset-based methods on Continual MetaWorld and Continual MinAtar benchmarks. Ablations reveal a tunable trade-off between plasticity and peak performance, highlighting utility-scaled reinitialization as a promising direction for continual learning.
}

\begin{document}

\maketitle  

\begin{abstract}
	Neural networks are hindered by accumulating dormant neurons and loss of expressivity throughout training, particularly in non-stationary data settings, such as continual supervised and reinforcement learning. Recently, neuron resets have been used to maintain gradient flow and restore plasticity. However, full unit reinitialization often sacrifices peak performance and can destabilize training, leading to policy collapse.
	To preserve plasticity without destabilizing training, we propose Calibrated Partial Resets (CPR), an optimizer that periodically pulls low-utility neurons toward their initialization, with pull strength scaled by each neuron's utility. Unlike binary reset methods, \emph{partial} resets avoid brittleness; unlike uniform decay, \emph{calibrated} utility-scaling concentrates adjustment on the units that need it most.
	Among compared methods, only CPR avoids policy collapse over 400M training steps in SlipperyAnt, and it outperforms prior decay and reset-based methods on Continual MetaWorld and Continual MinAtar benchmarks. Ablations reveal a tunable trade-off between plasticity and peak performance, highlighting utility-scaled reinitialization as a promising direction for continual learning.
\end{abstract}

\section{Introduction}
\label{sec:intro}
Deep neural networks perform well in stationary training regimes but often lose plasticity under prolonged non-stationary training, limiting continual supervised learning and reinforcement learning (RL) agents that must adapt over long horizons. In RL, plasticity loss is especially consequential because policy updates change the agent's future data distribution; small optimization instabilities can therefore compound through feedback loops into catastrophic policy degradation \citep{study_on-policy_rl_plasticity_loss}.

Prior work links plasticity loss to decreasing feature rank, rising parameter norms, and representation collapse \citep{lyle2022understanding}. \citet{redo} show that training distribution shifts cause an increase in previously active units becoming dormant, \citet{cbp_older,cbp} additionally show that decreases in stable rank correlate with reduced adaptation ability, and \citet{pmlr-v202-lyle23b} posit that unstable loss landscapes and optimization difficulties contribute to plasticity loss.

One remedy is to adaptively reinitialize low-utility neuron units during training, thereby restoring unused network capacity. Existing methods differ primarily in how they estimate utility. Continual Backpropagation (CBP) \citep{cbp_older,cbp} defines a unit’s utility as the running average of its activation weighted by its outgoing connections, and subsequently zeros those connections upon reinitialization. ReDo \citep{redo} uses batch-level activation statistics, while ReGraMa \citep{grama} uses gradient magnitude. Despite these differences, these methods share the same intervention structure: a unit below a threshold is fully reinitialized, while all other units are left unchanged.

We argue that this conflates two design choices that should be separated: which units should be refreshed, and how strongly they should be changed. Binary reset methods are selective but discontinuous, introducing abrupt parameter changes that can destabilize long-horizon RL training. Uniform decay methods, such as Shrink \& Perturb \citep{shrink_and_perturb}, make smoother changes but apply them indiscriminately across units. This exposes a missing design point: a method that is both selective and smooth.

To address this, we propose Calibrated Partial Resets (CPR), which periodically pulls neurons toward their initialization by a continuous coefficient determined by normalized per-neuron utility. Low-utility units are refreshed more strongly, while high-utility units are perturbed only weakly. Thus, CPR retains the selectivity of neuron-reset methods while avoiding the all-or-nothing updates that can make full reinitialization brittle in continual RL.


Our contributions can be summarized as follows:

\begin{itemize}
	\item We identify binary neuron resets as a source of instability in long-horizon continual RL and propose CPR, a utility-scaled partial reset optimizer.
	\item We show that CPR avoids policy collapse over 400M steps in SlipperyAnt and improves performance on SlipperyHumanoid, Continual MetaWorld, and Continual MinAtar.
	\item We ablate reset magnitude and transformation shape, showing that partial, utility-weighted resets control the plasticity–stability tradeoff.
\end{itemize}


\begin{figure}[tbp]
	\centering
	\begin{minipage}[b]{0.50\textwidth}
		\includegraphics[width=\linewidth,trim= 60 45 0 35, clip]{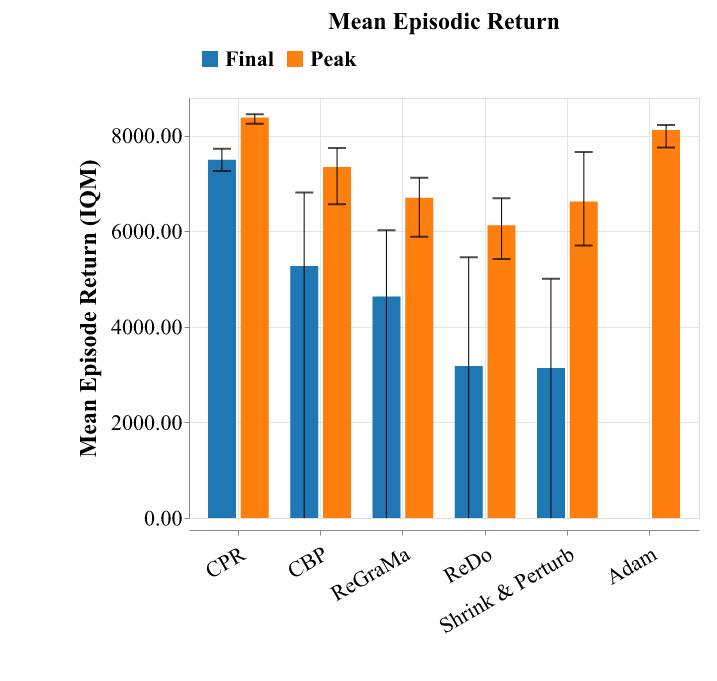}
	\end{minipage}
	\begin{minipage}[b]{0.45\textwidth}
		\centering
		\includegraphics[width=\linewidth, trim= 0 0 0 34,clip]{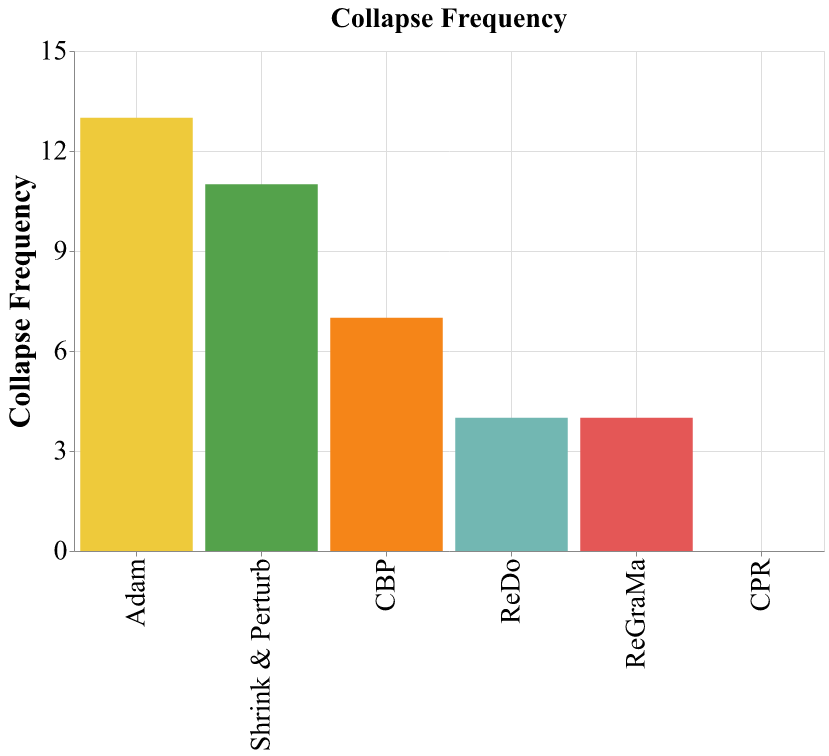}
	\end{minipage}
	\caption{\textbf{Left:} Peak vs. final Interquartile Mean (IQM) episodic return in SlipperyAnt with Interquartile Range (IQR) error bars. \textbf{Right}: number of seeds with policy collapses. CPR achieves the highest return and \textbf{is the only method with zero policy collapses across all 15 seeds over 400M steps} while Adam collapses and receives a mean episodic return below zero}
	\label{fig:flagship}

\end{figure}

\section{Background}
We consider the Continual RL setting, where an agent operates in a sequence of $N$ Markov decision processes (MDPs) $\{\mathcal{M}_1, \mathcal{M}_2, \ldots,\mathcal{M}_N\}$, each defined by the standard tuple $\mathcal{M}_k =(\mathcal{S}, \mathcal{A}, \mathcal{P}_k, \mathcal{R}_k, \gamma)$. The agent follows a stochastic policy $\pi : \mathcal{S} \times \mathcal{A} \rightarrow [0,1]$ and aims to maximize the discounted return $\mathbb{E}_{a_t \sim \pi} [\sum_{t=0}^{\infty} \gamma^t \mathcal{R}(s_t, a_t)]$. While RL agents face inherent non-stationary training dynamics in various forms even in standard MDPs \citep{igl2020transient}, we focus on problems with added non-stationarity imposed on the environment itself such that the agent must adapt over time. Given a training horizon $\mathcal{T}$, the environment switches to the next MDP every $\frac{\mathcal{T}}{N}$ timesteps, requiring the agent to continually adapt to changing dynamics.

\subsection{Network Plasticity}

We consider feed-forward networks with layers indexed by $\ell=1,\dots,L$. Let
$W^\ell\in\mathbb{R}^{n_\ell\times n_{\ell-1}}$ and $b^\ell\in\mathbb{R}^{n_\ell}$ denote the parameters that map
$h^{\ell-1}$ to pre-activations $z^\ell = W^\ell h^{\ell-1}+b^\ell$ and activations $h^\ell=\sigma(z^\ell)$.
By convention, the \emph{incoming} weights for unit $i$ at layer $\ell$ are the $i$-th \emph{row} of $W^\ell$
and the \emph{outgoing} weights are the $i$-th \emph{column} of $W^{\ell+1}$. We use mini-batches
$\mathcal{B}_t$ at update step $t$ and write $\mathbb{E}_{x\in \mathcal{B}_t}[\cdot]$ for batch averages.

\paragraph{Dormant-neuron ratio \citep{redo}.} A neuron is \textit{dormant} if its mean activation is small relative to the layer average.
\begin{equation}
	\label{eq:dormant_ratio}
	\mathrm{DormantRatio}_\ell \;=\; \frac{1}{n_\ell}\sum_{i=1}^{n_\ell}
	\mathbf{1}\!\left[\frac{\mathbb{E}_{x\sim \mathcal{B}_t}|h_i^\ell(x)|}{\frac{1}{n_\ell}\sum_{k=1}^{n_\ell}\mathbb{E}_{x\sim \mathcal{B}_t}|h_k^\ell(x)|} < \tau\right].
\end{equation}
where $\tau = 0.1$ in experiments.

\paragraph{Linearized-neuron ratio \citep{cause_of_loss}.}
We define the number of \emph{linearized units} per layer as the percentage of neurons for which the pre-activation value of a given neuron was positive. We can define linearization as:
\begin{equation}
	\label{eq:linearized_ratio}
	\mathrm{LinearizedRatio}_\ell =
	\frac{1}{n_\ell}\sum_{i=1}^{n_\ell}\mathbf{1}\!\left[\mathbb{E}_{x\in \mathcal{B}_t}[\mathbf{1}[\,z_i^\ell(x)>0\,]] > \theta\right]
\end{equation}
where $\theta=0.9$ in experiments.

These metrics complement each other: \autoref{eq:dormant_ratio} captures under-used channels, while \autoref{eq:linearized_ratio}
captures loss of gating diversity that reduces rank and gradient flow. We report both alongside accuracy/return. Note that these definitions are for rectifier-like activation functions only. For non-rectifier activations, an analogous variance-based criterion can be used \citep{cause_of_loss}.

\subsection{Unit Utility and Selective Reset}
A \emph{neuron-utility score} $S_i^\ell(t)$ quantifies how much unit $i$ at layer $\ell$ contributes to learning and can be computed given network weights $W^\ell$ and activations $h^\ell$ or averaged gradients $\mathbb{E}_{x\in \mathcal{B}_t}[\nabla_{W_i^\ell} \mathcal{L}(W^\ell)]$ where $\mathcal{L}$ is a loss function. To stabilize selection under non-stationarity, it is useful to compute a layer-normalized exponential moving average (EMA) of the score $u_i^\ell(t)$:
\begin{equation}
	\label{eq:ema}
	\begin{split}
		u_i^\ell(t) &\;\leftarrow\;\beta u_i^\ell(t-1) + (1-\beta)\,\,\tilde{S}_i^\ell(t),\qquad\\
		\tilde{S}_i^\ell(t) &\;=\;\frac{S_i^\ell(t)}{\frac{1}{n_\ell}\sum_{k=1}^{n_\ell} S_k^\ell(t)},\;\;\beta\in[0,1).
	\end{split}
\end{equation}

where $\beta$ is the EMA decay parameter. Normalization over layers (through dividing by the layer mean) makes selection thresholds comparable across layers and training time; see \autoref{fig:util_dist} for a visualization.

\paragraph{Selective reset methods.}
We study neuron-reset methods that operate immediately after the base optimizer update step and reinitialize units with low utility score $S_i^\ell$ or $u_i^\ell$ based on a \emph{reset rule} $\mathfrak{R}$.

\subsection{Training Instability in Continual Reinforcement Learning}
\label{sec:instabilityincrl}

Existing neuron-reset methods (e.g., CBP, ReDo, ReGraMa \citep{cbp, redo, grama}) differ in utility estimation and reset scheduling, but share a \emph{binary} intervention $\mathfrak{R}(S_i^\ell,\mathfrak{r})=\mathbf{1}[S_i^\ell(t) < \mathfrak{r}]$, which fully reinitializes low-utility neurons

\begin{equation}
	\label{eq:binary_reset}
	\begin{aligned}
		 & W^{\ell}_{i,:} \leftarrow (1 - \mathfrak{R}(S_i^\ell,\mathfrak{r}))\,W^{\ell}_{i,:} + \mathfrak{R}(S_i^\ell,\mathfrak{r})\xi^{\ell}_{i,:}\qquad \\
		 & W^{\ell+1}_{:,i}\leftarrow (1 - \mathfrak{R}(S_i^\ell,\mathfrak{r}))\,W^{\ell+1}_{:,i}
	\end{aligned}
\end{equation}

where $\xi^{\ell}_{i,:}\!\sim\!\mathcal{D}_{\text{init}}$, with $\mathcal{D}_{\text{init}}$ being a weight initialization distribution, for instance Kaiming \citep{kaiminginit} or Xavier \citep{xavierinit}. $W^{\ell}_{i,:}$ denotes the incoming weights into unit $i$ while $W^{\ell+1}_{:,i}$ are the outgoing weights.

Notably, under this reset rule, neurons either get reset if they are deemed low-utility, in which case their incoming weights are completely reinitialized, or they do not. This all-or-nothing reinitialization introduces sudden, large weight changes which are a known source of instability in deep RL \citep{overcoming_policy_collapse} and can cause gradient spikes that destabilize training. Additionally, units that are reinitialized have their outgoing weights zeroed, which can lead to a distribution shift in the inputs of the next layer, causing further training instability, especially in the case where layer normalization \citep{layernorm} is not used.

We empirically find that these instabilities can and do arise in these algorithms in continual RL tasks in \autoref{fig:grad_norm} and can also lead to policy collapse in \autoref{fig:flagship}. Typical policy collapse mitigation such as early stopping \citep{early_stopping} and gradient stabilization approaches such as learning rate annealing \citep{cosing_decay} are designed for stationary, single-task settings and cannot be readily applied to continual learning where ideally the agent should be able to learn throughout its lifetime up to an infinite horizon.

\section{Calibrated Partial Resets (CPR)}
\label{sec:method}

\paragraph{Utility score estimation.}

Following recent work showing gradient-based utility criteria scale
more favorably~\citep{grama, weightsvunits}, we define the raw
utility score for unit $i$ in layer $\ell$ as the mean gradient
magnitude of its incoming weights:
\begin{equation}
	\label{eq:CPR_utility}
	S_i^{\ell}(t) = \mathbb{E}_{x\in \mathcal{B}_t}\!\left[\,
		\left\|\nabla_{W_{i,:}^{\ell}} \mathcal{L}(x)\right\|\,\right]
\end{equation}
where $\mathcal{L}$ is the loss function being optimized. This score
is then layer-normalized and smoothed via the EMA defined in
\autoref{eq:ema} to obtain $u_i^\ell$.

\paragraph{Reset Mechanism.}
CPR differs from prior work along two axes. Binary reset methods (\autoref{sec:instabilityincrl}) fully reinitialize selected units, which introduces abrupt parameter changes at reset events. Decay-based methods such as Shrink \& Perturb \citep{shrink_and_perturb} apply a uniform shrinkage to all units, ignoring per-neuron utility. CPR is \emph{partial} on one axis and \emph{calibrated} on the other: every $f$ steps, each unit's weights are pulled toward initialization by a coefficient $r_i^\ell = \rho \phi(u_i^\ell) \in [0, \rho]$, where $\phi$ is a monotonically decreasing function of the unit's utility and $\rho \in (0,1]$. We use \emph{calibrated} to emphasize that the reset magnitude is set per-neuron from a measured signal, rather than applied uniformly (decay) or as an all-or-nothing event (binary reset).

The two design choices contribute roughly independently. Avoiding full reinitialization reduces the gradient spikes that destabilize binary-reset methods, regardless of how aggressively low-utility units are targeted. Targeting via utility even coarsely, e.g. by resetting the below-mean-utility units in each layer, strongly outperforms uniform shrinkage. We map utility to a reset intensity via the shape function
\begin{equation}
	\label{eq:util_fn}
	\phi(u_i^\ell) = \min\!\left(2\sigma\!\left[-\kappa(u_i^\ell-1)\right],\,1\right),
\end{equation}
where $u_i^\ell$ is the per-layer normalized utility with mean $1$; $\sigma$ is the logistic sigmoid function $\sigma(x) = (1+e^{-x})^{-1}$; and $\kappa > 0$ determines the sharpness of the mapping. The factor of 2 anchors the curve at the layer mean ($\phi(u_i^\ell)=1$ when $u_i^\ell=1$). Units with above average utility receive progressively smaller resets, while those below the mean receive the maximum per-unit reset fraction $\rho$ (alternative smooth shapes have a measurable but secondary effect; see \autoref{Appendix:transformations} and \autoref{fig:util_comp}).

Notably, as $\kappa \to \infty$, $\phi$ approaches a step function: $\lim_{\kappa \to\infty}\phi(u) = \mathbf{1}[\,u \le 1\,]$, applying resets only to units below the layer-average utility. Even in this limit, CPR remains partial: the below-mean-utility units in each layer are pulled toward initialization by $\rho$ rather than fully reinitialized. This isolates the contribution of partial-vs-full reset from that of utility-based targeting. At the opposite extreme, $\kappa=0$ applies uniform resets to all neurons. Empirically, our sharpness ablation (\autoref{fig:sharpness_ablation}) shows that CPR is robust across a range of $\kappa \in [2,20]$, with performance degrading only at small $\kappa \leq 1$ values where the mapping becomes nearly uniform and the per-neuron signal is lost. Given this wide plateau, we fix $\kappa=16$ across all experiments rather than treating it as a tunable hyperparameter.

The resulting partial reset operator for unit $i$ in layer $\ell$ is
\begin{equation}
	\label{eq:weight_reset}
	\begin{aligned}
		 & W^{\ell}_{i,:} \leftarrow (1 - r_i^\ell)\,W^{\ell}_{i,:} + r_i^\ell\, \xi^{\ell}_{i,:},\qquad \\
		 & W^{\ell+1}_{:,i} \leftarrow (1 - r_i^\ell)\,W^{\ell+1}_{:,i},
	\end{aligned}
\end{equation}
where, as in \autoref{eq:binary_reset}, $\xi^{\ell}_{i,:}\!\sim\!\mathcal{D}_{\text{init}}$. We perform CPR updates every update frequency $f$ and compute the EMA of utility scores between update steps only. This is implemented by using data from every optimization step to compute the utility score EMA, but along with CPR updates, we reset utility running averages to their mean $1$. Algorithm~\autoref{alg:CPR} in \autoref{sec:cpr_algo} gives the complete update algorithm, including utility-EMA accumulation, reset scheduling and utility re-centering.

\section{Results}
\paragraph{Experiment setup.}
\label{sec:expr_setup}
Our benchmark suite includes long-horizon friction-shift control tasks (SlipperyAnt and SlipperyHumanoid), continual off-policy control (Continual MetaWorld), and continual visual control (Continual MinAtar), with appendix experiments in single-task RL and continual supervised learning.

We evaluate CPR against recent continual learning methods in two challenging continual RL settings: SlipperyAnt and SlipperyHumanoid. These are modified Brax-based variants of the OpenAI Gym Ant and Humanoid environments \citep{openai_gym, brax}, similar to the continual RL setting studied by \citet{cbp}. SlipperyAnt\footnote{While SlipperyAnt was studied in prior work \citep{cbp}, implementing it on top of the Brax version leads to different rewards and dynamics.} requires a running ant-like robot to adapt to global friction changes, while SlipperyHumanoid requires a running humanoid to adapt as ground friction changes periodically. This challenges continual learning agents, as they must retain locomotion ability while selectively forgetting friction dynamics in their implicit world model.

We train agents for 400 million steps, sample friction log‑uniformly from $[0.02, 2.0]$ every 20 million steps and use 2048 parallel environments. We additionally open source the codebase for these experiments\footnote{github.com/LucMc/continual-learning/}.

As the base RL algorithm we use Proximal Policy Optimization (PPO) \citep{ppo} with Adam \citep{adam} as the base optimizer, with hyperparameters and network sizes provided in \autoref{Appendix:hyperparameter}.

SlipperyAnt and SlipperyHumanoid use 15 random seeds; MT1 uses 10 seeds with Continual MetaWorld and Continual MinAtar using 5 seeds. We provide further details for these benchmarks in \autoref{Appendix:impl}. We report the IQM across seeds, as proposed by \citet{rliable}. Interquartile ranges (IQR) are provided in \autoref{Appendix:more_results} (\Cref{tab:ant_iqr,tab:hum_iqr}). We selected the hyperparameters with highest average final return (\autoref{Appendix:hyperparameter}, \autoref{tab:hyper_sweep}).

\subsection{CPR Maintains or Improves Peak Performance}
\begin{figure}[t]
	\centering
	\begin{minipage}{0.49\linewidth} 
		\includegraphics[width=\linewidth]{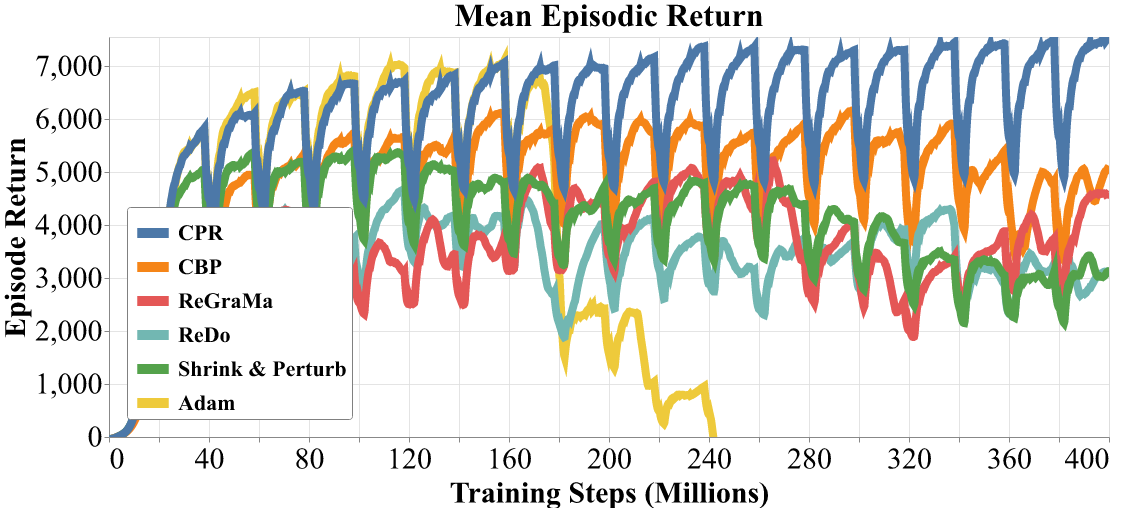}
	\end{minipage}
	\begin{minipage}{0.49\linewidth} 
		\includegraphics[width=\linewidth]{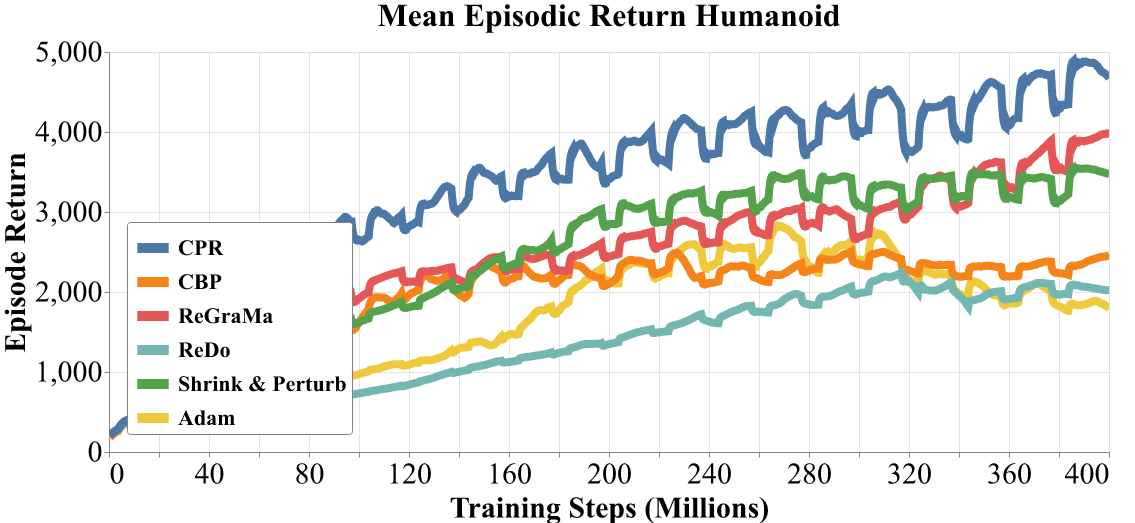}
	\end{minipage}

	\caption{Comparison of IQM of episodic returns over 15 seeds. \textbf{Left}: SlipperyAnt. \textbf{Right}: SlipperyHumanoid. IQR is removed for visual clarity and can be found in \Cref{tab:ant_iqr,tab:hum_iqr}. CPR maintains high return in SlipperyAnt while leading throughout training in SlipperyHumanoid.}
	\label{fig:ant_humanoid_main}
\end{figure}

On SlipperyAnt, we observe in \autoref{fig:ant_humanoid_main} that CPR can consistently achieve peak performance after each friction change. CPR maintains this ability even after 400 million steps of training. Adam with no neuron-reset method attached achieves strong peak performance but eventually collapses. Meanwhile, other reset-based methods such as ReDo and ReGraMa with a binary reset rule peak later on in training before themselves showing a decline in performance. This illustrates a plasticity-performance trade-off in this setting where existing methods either peak high and collapse, such as Adam which has no mechanism to maintain plasticity, or maintain trainability but are unable to learn optimal policies for each task due to unstable training dynamics arising from the neuron resets that help maintain gradient flow. In contrast, CPR can achieve both objectives.

In SlipperyHumanoid (\autoref{fig:ant_humanoid_main}), CPR substantially outperforms baselines in mean episodic return throughout training. This suggests more effective transfer across task changes.

We evaluate CPR in two further Continual RL benchmarks inspired by \citet{continualworld} and \citet{continualminatar}, using Continual MetaWorld and Continual MinAtar respectively. Continual MetaWorld sequentially iterates through the MT10 taskset using SAC and Muon as the base optimizer. Continual MinAtar evaluates methods on a sequence of Atari tasks using a CNN architecture.

\begin{table}[h]
	\centering
	\begin{minipage}[t]{0.48\linewidth}
		\centering
		\caption{Continual MetaWorld IQM of return over training steps and final performance using 5 seeds with IQR}
		\label{tab:continual_metaworld}
		\small
		\setlength{\tabcolsep}{4pt}
		\begin{tabular}{lcc}
			\toprule
			Method            & Avg                & Final             \\
			\midrule
			CPR               & \textbf{0.231 $\pm$ 0.005}  & \textbf{0.200 $\pm$ 0.010} \\
			CBP               & 0.204  $\pm$ 0.009 & 0.100 $\pm$ 0.030 \\
			ReDo              & 0.199 $\pm$ 0.017  & 0.137 $\pm$ 0.025 \\
			ReGraMa           & 0.196 $\pm$ 0.003  & 0.155 $\pm$ 0.007 \\
			Shrink \& Perturb & 0.199 $\pm$ 0.007  & 0.150 $\pm$ 0.025 \\
			Muon              & 0.179 $\pm$ 0.008  & 0.145 $\pm$ 0.028 \\
			Adam              & 0.147 $\pm$ 0.003  & 0.167 $\pm$ 0.023 \\
		\end{tabular}
	\end{minipage}\hfill
	\begin{minipage}[t]{0.48\linewidth}
		\centering
		\caption{Continual MinAtar IQM of return over training steps and final performance using 5 seeds with IQR}
		\label{tab:continual_minatar}
		\small
		\setlength{\tabcolsep}{4pt}
		\begin{tabular}{lcc}
			\toprule
			Method            & Avg             & Final           \\
			\midrule
			CPR               & \textbf{75.6 $\pm$ 14.3} & \textbf{52.3 $\pm$ 25.0} \\
			CBP               & 65.6 $\pm$ 3.2  & 5.8 $\pm$ 14.9  \\
			ReDo              & 68.0 $\pm$ 7.6  & 11.5 $\pm$ 11.7 \\
			ReGraMa           & 65.5 $\pm$ 2.1  & 15.0 $\pm$ 13.6 \\
			Shrink \& Perturb & 61.1 $\pm$ 8.1  & 3.3 $\pm$ 2.6   \\
			Adam              & 54.9 $\pm$ 3.5  & 1.9 $\pm$ 0.8   \\
		\end{tabular}
	\end{minipage}
\end{table}

\subsection{CPR Improves Training Stability}

\begin{figure}[t]
	\centering
	\begin{minipage}[t]{0.49\linewidth}
		\centering
		\includegraphics[width=\linewidth]{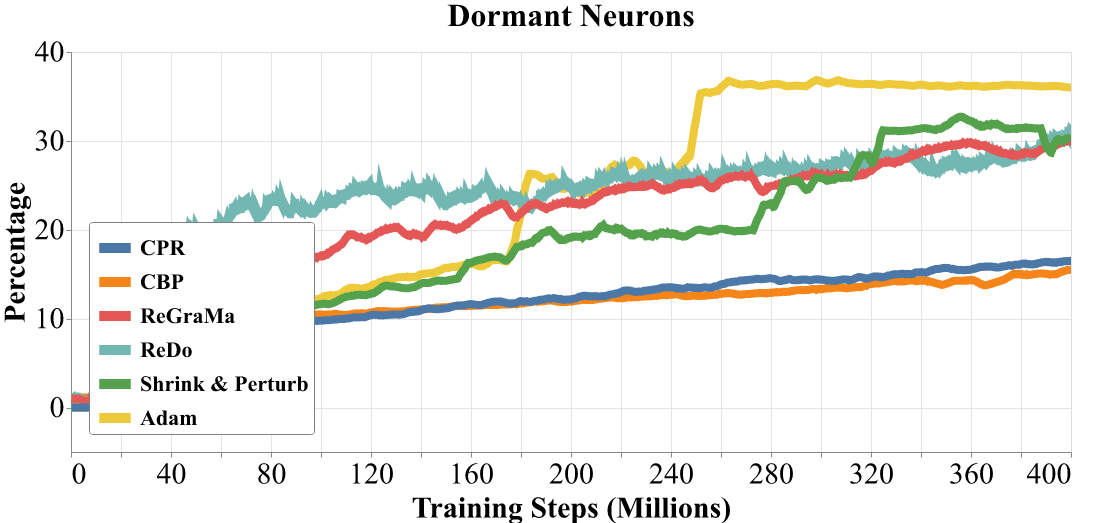}
	\end{minipage}\hfill
	\begin{minipage}[t]{0.49\linewidth}
		\centering
		\includegraphics[width=\linewidth]{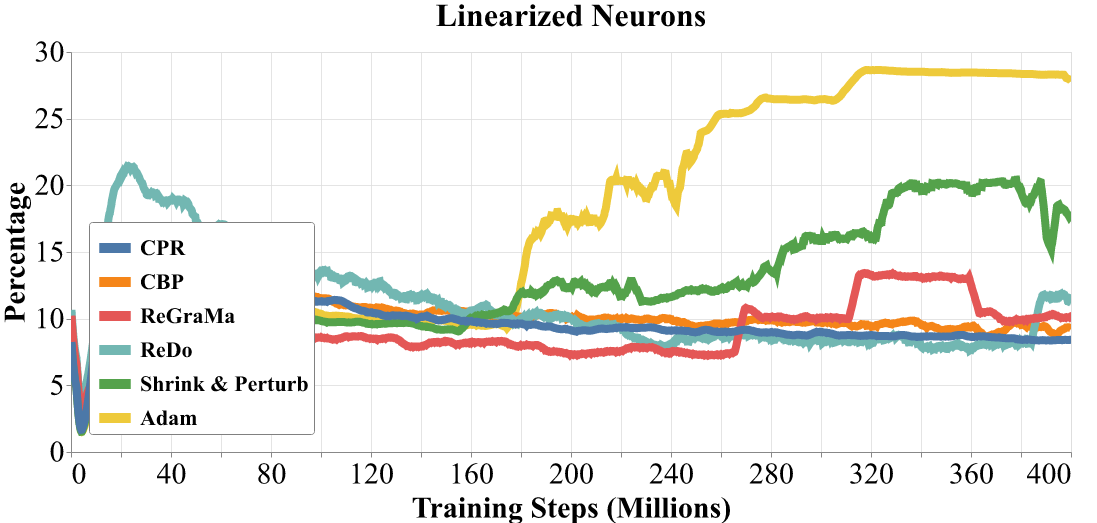}
	\end{minipage}

	\caption{\textbf{Left:} Dormant‑neuron ratio. \textbf{Right:} Linearized-neuron ratio. IQM across 15 seeds (lower is better). Lower dormancy correlates with higher plasticity and lower linearization correlates with higher representation capacity. This shows all methods reduce unit dormancy and linearization, though CBP and CPR appear most effective. }
	\label{fig:dormant_linearized}
\end{figure}

\paragraph{Dormant and linearized neurons.}
\autoref{fig:dormant_linearized} shows that both CPR and CBP keep the dormant‑neuron ratio nearly flat over 400M steps, whereas Adam and baselines that don't use running statistics of the utility score rise steadily, with acceleration after $\sim$200–300M steps, which coincides with the return degradation in \autoref{fig:ant_humanoid_main}. Methods that strongly suppress dormancy tend to have higher linearized‑unit ratios initially; CPR and CBP maintain steady dormancy and linearization ratios, whereas other methods oscillate. Thus, CPR preserves plasticity-related statistics without the large late-training drift observed in several baselines.

\paragraph{Gradient norm.}
\autoref{fig:grad_norm} reports actor gradient norms for SlipperyAnt, with SlipperyHumanoid gradient norms shown in \autoref{fig:humanoid_grad_norm}. CPR exhibits smaller fluctuations over time relative to baselines with binary reset rules, which aligns with its ability to avoid collapses and sustain high final return. Parameter-norm dynamics show a similar pattern (\autoref{fig:param_norm}): baseline methods exhibit higher parameter growth, which \citet{cbp} identify as a key driver of plasticity loss, while CPR constrains parameter norms throughout training.

\subsection{CPR Prevents Policy Collapse}
\label{sec:policycollapse}
\begin{figure}[t]
        \vspace{0pt}
	\begin{minipage}{0.49\textwidth}
		\includegraphics[height=3.5cm, trim=0 0 0 50, clip]{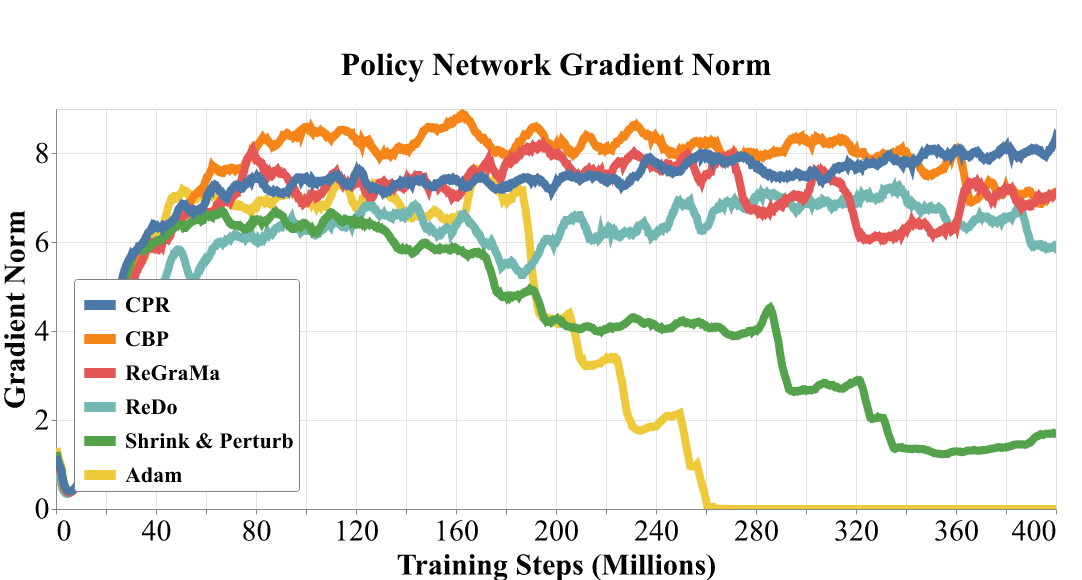}
		\caption{Actor gradient norm IQM over 15 seeds in SlipperyAnt}
		\label{fig:grad_norm}
	\end{minipage}
	\hfill
	\begin{minipage}{0.485\textwidth}
		\includegraphics[height=4.45cm, trim=0 0 20 -110,clip]{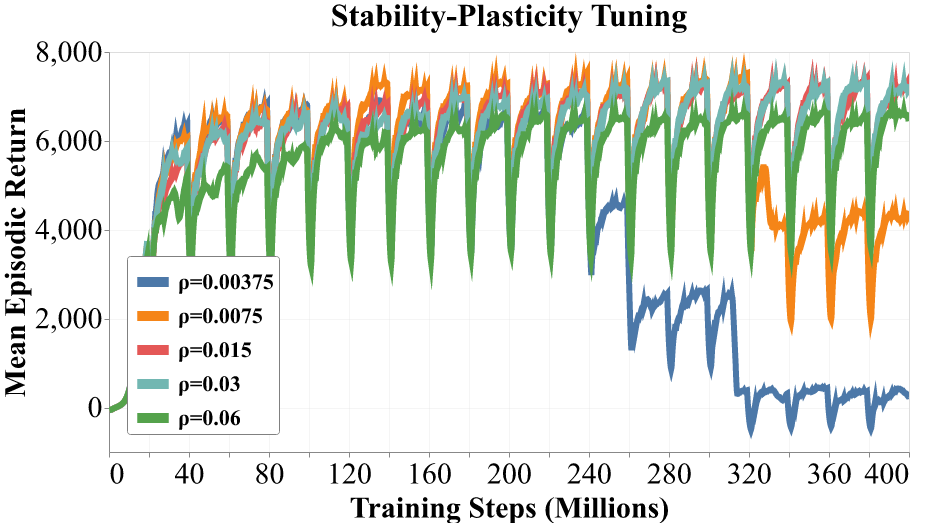}
		\caption{Tuning ablation for max per-unit reset $\rho$ in SlipperyAnt. A low $\rho$ preserves stability but hinders adaptation, while high $\rho$ values enhance plasticity but suppress peak reward}
		\label{fig:rr}
	\end{minipage}
\end{figure}

SlipperyAnt's 400M-step horizon and periodic friction shifts make it our primary stress test for training stability: it is the setting in which the instabilities of binary reset methods manifest most visibly. We use it as a diagnostic for catastrophic policy degradation; the broader stability benefits of CPR, reflected in gradient and parameter norm dynamics (\autoref{fig:grad_norm}, \autoref{fig:param_norm}) and in average performance on Continual MetaWorld and Continual MinAtar (\autoref{tab:continual_metaworld}, \autoref{tab:continual_minatar}), extend beyond this setting.

In SlipperyAnt, we measure a policy collapse as a sustained drop of at least $8{,}000$ in episodic return from a prior peak, persisting for at least 4M timesteps. The $8{,}000$ threshold spans the dynamic range of the environment (each method achieves a peak return of at least $\approx 6{,}000$, and episodic return is lower-bounded near $-2{,}000$); smaller thresholds risk classifying transient drops at task boundaries as collapses, while larger ones exclude methods whose peaks do not exceed the gap. The 4M-step persistence requirement excludes single-update spikes from natural variance.

The separation between CPR and reset-based baselines is not an artifact of this choice. Under a stricter $4{,}000$ threshold, CPR remains the only method with zero collapses across all 15 seeds, while every binary-reset baseline and Adam exhibit collapses on more than 40\% of runs. Under a lenient $10{,}000$ threshold, CPR, ReDo, and ReGraMa each have zero collapses, but this threshold excludes any method whose peak return does not exceed it. CPR uniquely avoids collapse across the meaningful range of the threshold (\autoref{fig:flagship}).

\paragraph{Controlling the stability--plasticity tradeoff.}
In CPR, $\rho$ acts as a direct control over the stability--plasticity balance. Small values of $\rho$ minimize the disruption caused by each reset, but in challenging environments such as SlipperyAnt, they may fail to inject sufficient plasticity for sustained adaptation, leading to gradual performance degradation.

Conversely, large values of $\rho$ increase plasticity, but can reduce peak and average performance by over-resetting useful structure.
Despite this tradeoff, our sweeps reveal a robust operational range of $\rho \in [0.01, 0.05]$ in which CPR consistently prevents policy collapse across both SlipperyAnt and SlipperyHumanoid.
Thus, while $\rho$ can be tuned to optimize the stability--plasticity tradeoff for a given environment, collapse prevention is not brittle to precise hyperparameter selection.
We empirically demonstrate the effect of ablating this parameter in SlipperyAnt in \autoref{fig:rr}.

%
%
%
\section{Discussion}
Calibrated Partial Resets (CPR) fills the gap between uniform decay and binary neuron reinitialization: it preserves the selectivity of reset methods while replacing all-or-nothing updates with utility-scaled partial resets. Across SlipperyAnt, SlipperyHumanoid, Continual MetaWorld, and Continual MinAtar, CPR improves long-horizon continual RL stability and performance. In the 400M-step friction-shift setting, CPR is the only method to avoid policy collapse while recovering peak performance after task changes. Although the permuted MNIST experiments use more tasks than any other, the difficulty is low enough that all methods except Adam saturate the benchmark. Similarly, Continual MinAtar (\autoref{tab:continual_minatar}) experiences less plasticity loss than SlipperyAnt as it contains only 2 task transitions, highlighting the difficulty in designing and assessing plasticity loss benchmarks.

We attribute CPR's collapse avoidance to utility-scaled partial resets, which inject plasticity without the sharp parameter updates induced by full reinitialization. This absence of collapse is a qualitative shift in training reliability, especially for long-horizon RL deployments where collapse may be difficult to recover from. Gradient, dormant-unit, linearization, and parameter-norm diagnostics provide convergent evidence for this mechanism, though no single diagnostic is decisive. The maximum per-unit reset fraction $\rho$ provides direct control over the plasticity--performance tradeoff.


\paragraph{Limitations.}
\label{para:limitations}
While we observe a substantial improvement in training stability and performance in the continual RL setting, CPR introduces implementation complexity compared to simpler heuristics such as Shrink \& Perturb. However, this is largely mitigated by our modular Optax implementation, which requires minimal modification to standard training pipelines. Additionally, calculating and storing per-neuron utility scores incurs a marginal computational overhead comparable to baselines; in our highest-dimensional setting, SlipperyHumanoid, we measured a $\sim 6\%$ increase in wall-clock training time compared to Adam (see \autoref{tab:runtime}). Finally, CPR introduces three tunable hyperparameters, the same as baselines. For the maximum per-unit reset fraction $\rho$ and decay rate $\beta$, we found that defaults of $\rho=0.015$ and $\beta=0.99$ generalize across our continuous control suite without task-specific tuning, though modest gains can be made by adjusting $\rho$ per environment.

Our experiments focus primarily on non-stationarity induced by changing task dynamics and task identity. We do not claim that CPR resolves all forms of continual RL instability, such as reward non-stationarity, observation shifts, morphology changes, or large-scale visual control. Similarly, our diagnostics support but do not fully identify the causal pathway by which partial resets prevent collapse. Future work should isolate roles of utility estimation, partial reinitialization, optimizer-state handling, and normalization under a wider range of architectures and non-stationary processes.


\bibliography{main}
\bibliographystyle{rlj}

\appendix

\section{CPR Algorithm}
\label{sec:cpr_algo}

\begin{algorithm}[htbp!]
	\caption{CPR (utilities every step; resets every $f$ steps)}
	\label{alg:CPR}
	\begin{algorithmic}[1]
		\REQUIRE Update frequency $f$, EMA decay $\beta$, steepness $\kappa$, max reset $\rho$
		\STATE \textbf{Input:} current step $t$
		\STATE Compute per-neuron utility $S_i^\ell(t)$  \COMMENT{layer-normalized}
		\STATE Update EMA: $u_i^\ell \gets \beta\,u_i^\ell + (1-\beta)\,\tilde{S}_i^\ell$
		\IF[apply resets every $f$ steps]{$t \bmod f = 0$ and $t > 0$}
		\FOR{each layer $\ell$}
		\FOR{each neuron $i$}
		\STATE $r_i^\ell \gets \rho \cdot\min\!\left(\,2 \times \sigma\!\big(-\kappa (u_i^\ell-1)\big),\,1\right)$ \COMMENT{See Equation \ref{eq:util_fn}}
		\STATE $W_{i,:}^{\ell} \gets (1-r_i^\ell)\,W_{i,:}^{\ell} + r_i^\ell\,\xi^{\ell}_{i,:}$
		\STATE $W_{:,i}^{\ell+1} \gets (1-r_i^\ell)\,W_{:,i}^{\ell+1}$
		\STATE $u_i^\ell \gets 1$ \COMMENT{reset running utility after a reset step}
		\ENDFOR
		\ENDFOR
		\ENDIF
	\end{algorithmic}
\end{algorithm}

\section{Additional Related Work}


\paragraph{Plasticity Loss in Deep Learning and RL.}
Training stability is a central challenge in RL, arising from function approximation, bootstrapping, and off-policy updates \citep{rl_intro, rl_triad_explained}. Policy shifts can be amplified through feedback loops, spiraling into catastrophic performance collapse \citep{study_on-policy_rl_plasticity_loss}. Accordingly, many major advances in RL have centered on improving stability including replay buffers \citep{dqn} and trust regions \citep{trpo}. Neural networks under prolonged non-stationary training forget prior knowledge \citep{catastrophic_forgetting,overcoming_forgetting} and lose their ability to adapt \citep{cause_of_loss,overestimation, cbp_older}, a phenomenon that has drawn increasing attention across class-incremental learning \citep{classincremental}, supervised learning \citep{shrink_and_perturb}, and RL \citep{lyle2022understanding}. \citet{cbp} links plasticity loss to rising parameter norms, dormant units, and reductions in stable rank. These findings motivate methods that restore trainability by controlling weight statistics or re-introducing randomness, including L2 regularization \citep{kumar2023maintainingL2}, shrink \& perturb \citep{shrink_and_perturb}, and binary reset methods \citep{redo,cbp,grama}.

\paragraph{Reinitialization Algorithms.} A closer line of work preserves plasticity by periodically reinitializing low-utility units, motivated by work that demonstrates how plasticity loss can be attributed to only a subset of neurons that become dormant \citep{cbp_older,redo}. Normalization \citep{layernorm} can mitigate plasticity loss, but prior work \citet{weightsvunits} suggests reinitialization provides additional benefit. CBP \citep{cbp_older,cbp} identifies inactive neurons by using moving averages of activation statistics and continually resets them. More recent work such as ReDo \citep{redo} initializes dormant neurons periodically and uses activation statistics computed over a single batch of data, simplifying implementation over CBP without cost to performance. Alternative approaches such as SNR \citep{SNR} count the time between feature activations and ReGraMa \citep{grama} uses gradient magnitudes as a more scalable metric for unused neurons than activation statistics.

CPR uses gradient magnitude utilities, as in ReGraMa, but smooths them over multiple optimization steps, as in CBP. Additionally, unlike prior reset methods, CPR does not fully reinitialize a targeted subset of neurons, but instead partially reinitializes all neurons according to their utility. Our approach therefore sits between the uniform perturbations of decay-based methods \citep{shrink_and_perturb} and binary reinitialization of neuron reset methods. Our 400M-step experiments extend beyond typical horizons in prior reset-method evaluations and expose long-horizon instabilities in binary reset methods that CPR avoids.

\section{Experiments in Standard RL}
\label{Appendix:metaworld}

To assess whether the plasticity mechanisms in CPR induce instability or performance drops in stationary environments, we evaluate the method on the MetaWorld MT1 benchmark \citep{mclean2025meta} using Soft Actor Critic (SAC) \citep{haarnoja2018soft}. We chose SAC over PPO as it is standard for the benchmark and allows us to verify CPR's performance under off-policy methods. We compare CPR against Adam and other reset-based baselines across 10 distinct manipulation tasks over 2 million training steps. Reset methods hyperparameters are reported in \autoref{tab:mt1_reset_params}. We use the same SAC configuration as in \autoref{tab:cmw_sac_params}, except that the single-task MT1 experiments use a 1M replay buffer and replay ratio 4; all methods are evaluated under the same MT1 settings.

\subsection{Analysis of Results}
\begin{figure*}[htbp]
	\centering
	\begin{subfigure}{0.44\textwidth}
		\includegraphics[width=\linewidth]{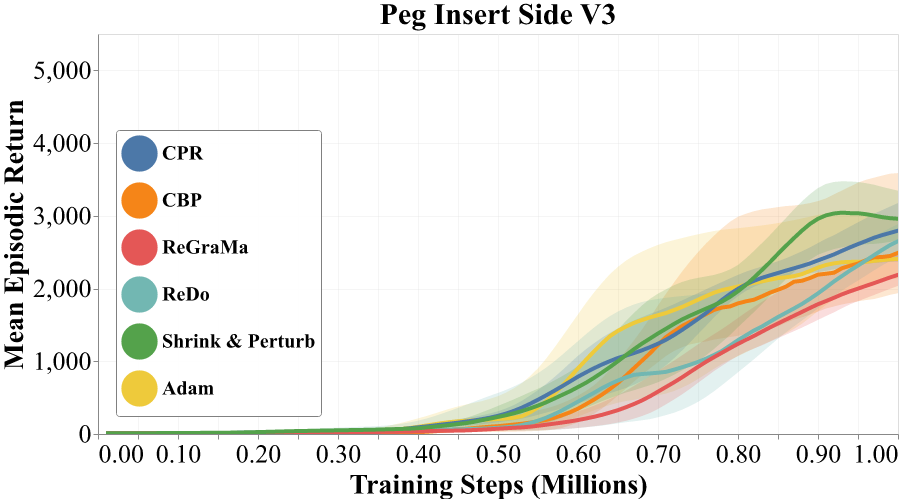}
	\end{subfigure}
	\hfill
	\begin{subfigure}{0.44\textwidth}
		\includegraphics[width=\linewidth]{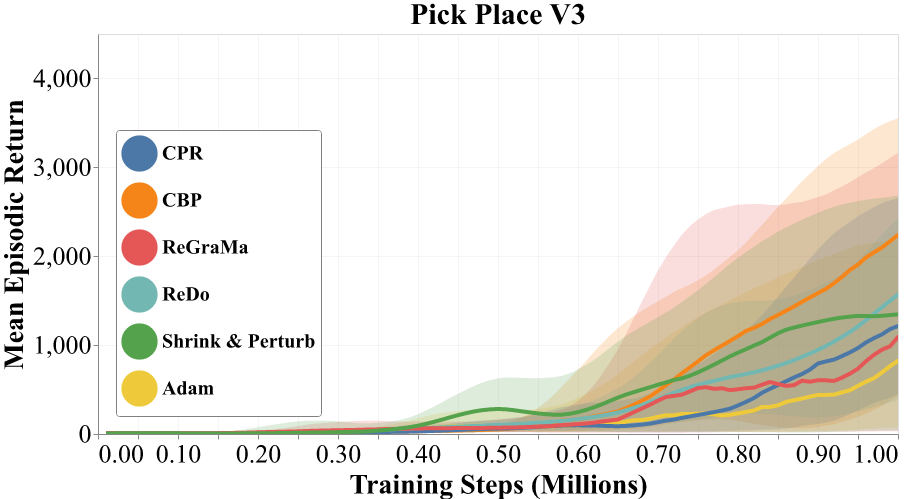}
	\end{subfigure}

	\vspace{1em}

	\begin{subfigure}{0.44\textwidth}
		\includegraphics[width=\linewidth]{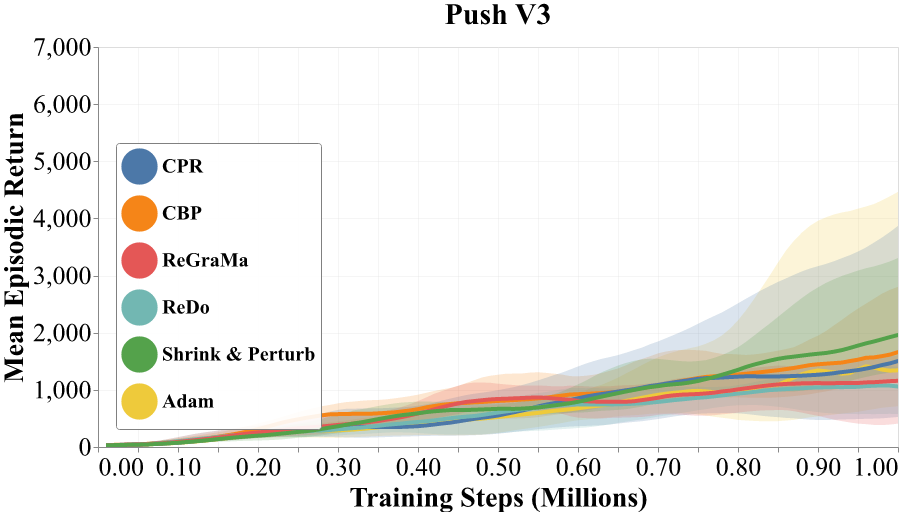}
	\end{subfigure}
	\hfill
	\begin{subfigure}{0.44\textwidth}
		\includegraphics[width=\linewidth]{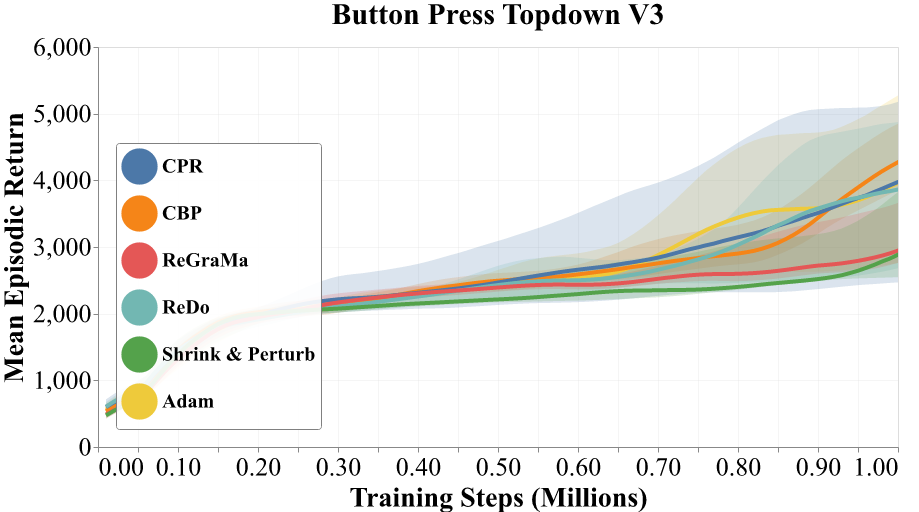}
	\end{subfigure}

	\vspace{1em}

	\begin{subfigure}{0.44\textwidth}
		\includegraphics[width=\linewidth]{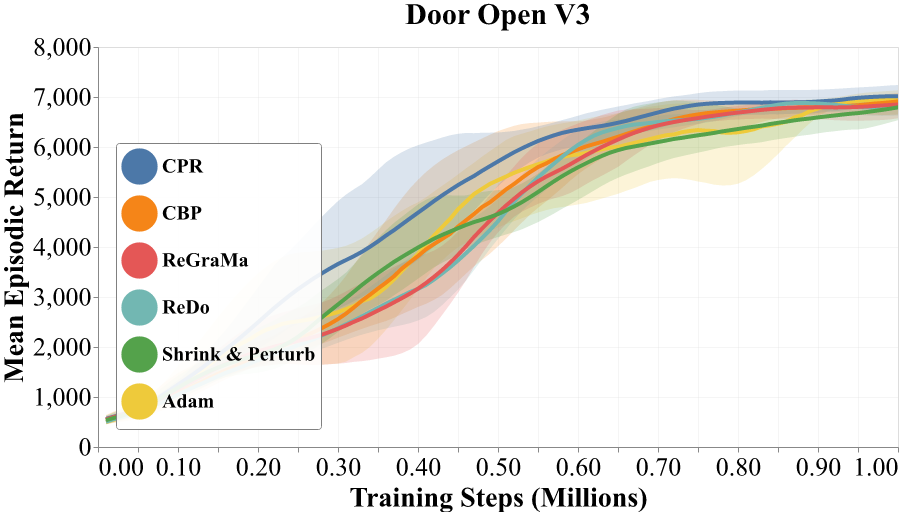}
	\end{subfigure}
	\hfill
	\begin{subfigure}{0.44\textwidth}
		\includegraphics[width=\linewidth]{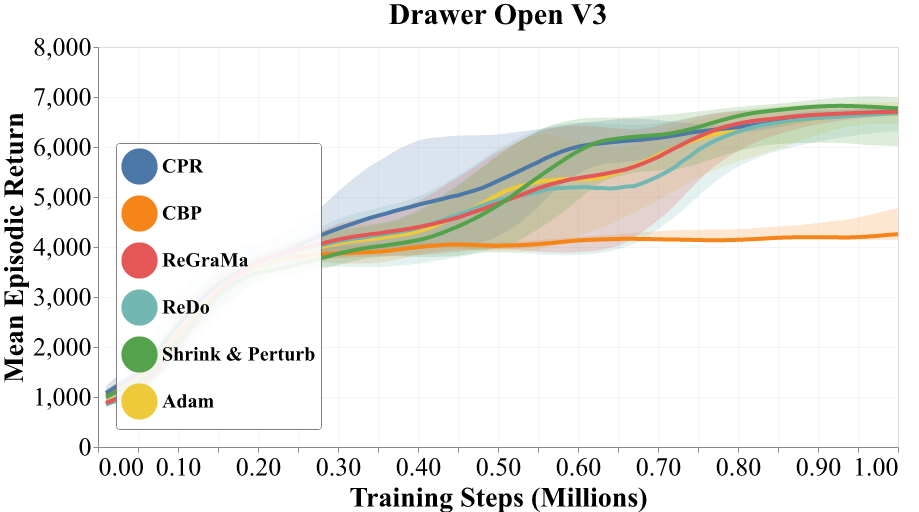}
	\end{subfigure}

	\vspace{1em}

	\begin{subfigure}{0.44\textwidth}
		\includegraphics[width=\linewidth]{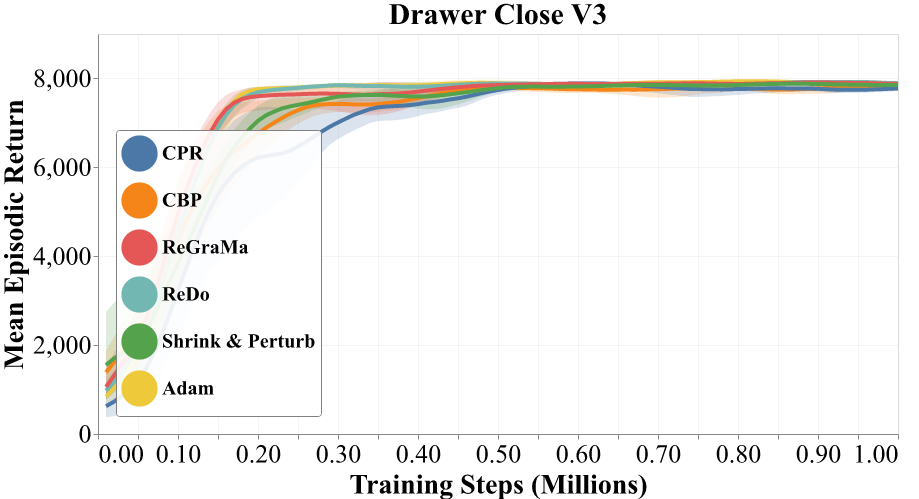}
	\end{subfigure}
	\hfill
	\begin{subfigure}{0.44\textwidth}
		\includegraphics[width=\linewidth]{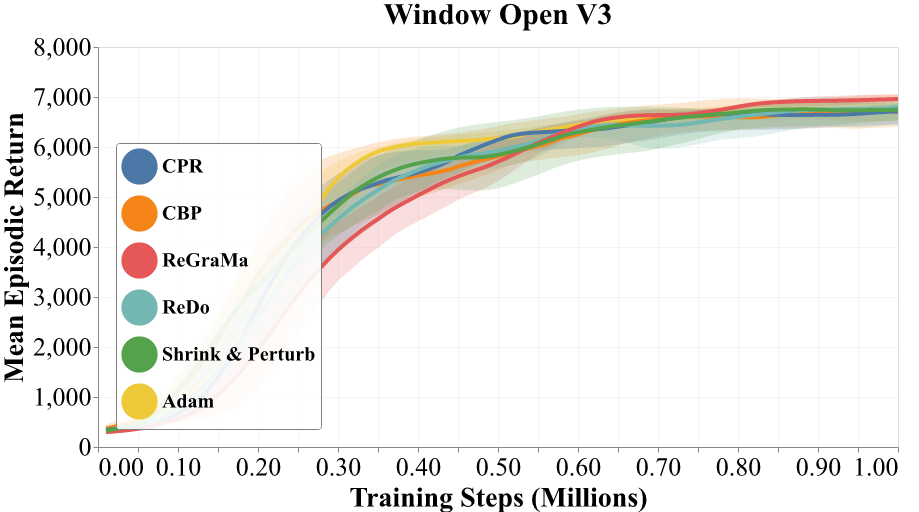}
	\end{subfigure}

	\vspace{1em}

	\begin{subfigure}{0.44\textwidth}
		\includegraphics[width=\linewidth]{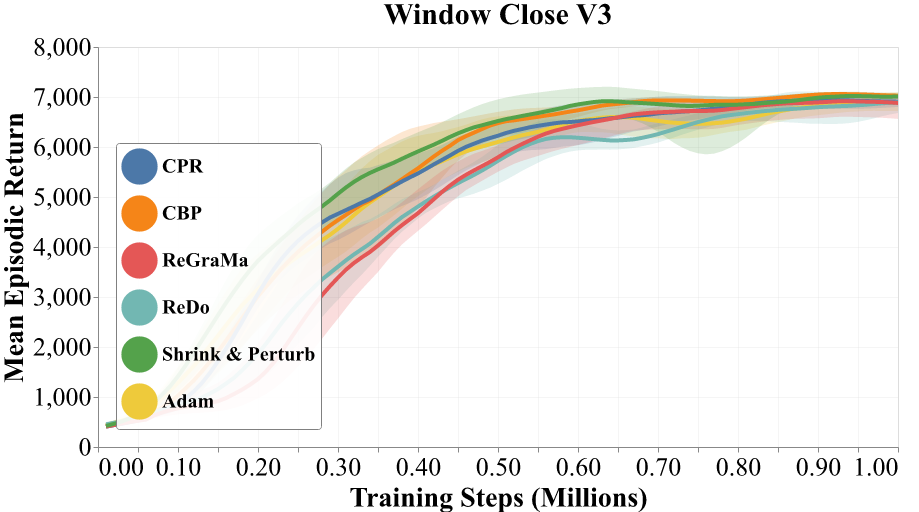}
	\end{subfigure}
	\hfill
	\begin{subfigure}{0.44\textwidth}
		\includegraphics[width=\linewidth]{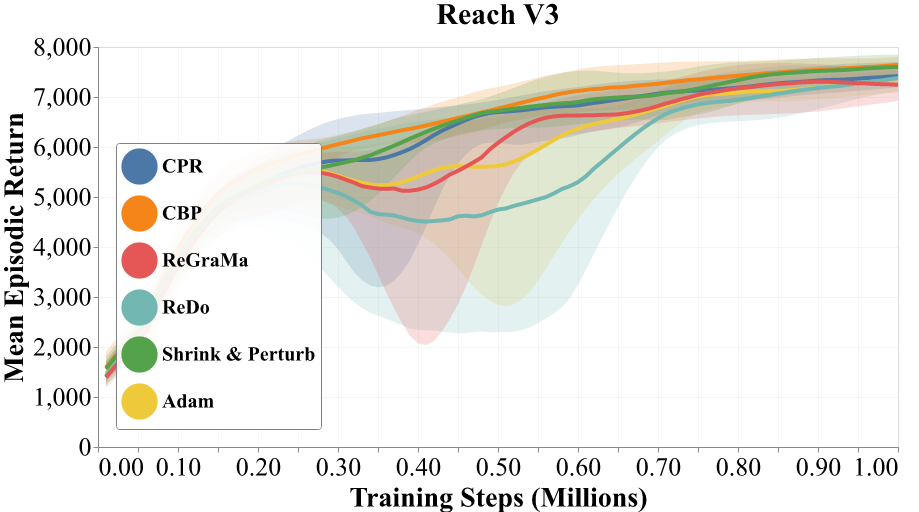}
	\end{subfigure}

	\caption{Mean episodic return over 2M steps for 10 MetaWorld continuous-control tasks (IQM over 10 seeds, shaded area indicates IQR)}
	\label{fig:metaworld_grid_2x5}
\end{figure*}

The results, visualized in Figure~\ref{fig:metaworld_grid_2x5}, demonstrate that CPR generalizes effectively to stationary settings, imposing no performance penalty compared to the base optimizer. We observe a saturation in performance across methods, particularly on easier tasks: the 2M-step horizon is insufficient to induce catastrophic plasticity loss under standard Adam, so baselines appear similarly capable in this regime. The differentiating benefits of CPR emerge primarily once task boundaries force rapid distribution shifts (Continual MetaWorld, \autoref{tab:continual_metaworld}) or training horizons extend into the regime where plasticity loss compounds into collapse (SlipperyAnt, 400M steps). We include MT1 results not as the principal evidence for CPR, but to verify the method imposes no penalty in stationary settings, a necessary property for a general-purpose continual learning optimizer.

\paragraph{Robustness and sample efficiency.} On challenging tasks such as Peg Insert Side and Pick Place, CPR tracks the performance profile of standard Adam or exceeds it (\autoref{fig:metaworld_grid_2x5}). This suggests that calibrated partial resets maintain the network's effective rank without disrupting the policy during the critical exploration phases of training. Reset-based baselines like ReDo, which improve plasticity in the continual setting, occasionally exhibit minor performance drops here (e.g., Reach-v3), while CPR matches the stability of Adam, retaining the benefits of plasticity without the volatility of binary resets.

\section{Experiments in Continual Supervised Learning}
\label{Appendix:cl}

\begin{figure}[!htbp]
	\centering
	\includegraphics[width=0.8\linewidth, trim=0 30 32 0, clip]{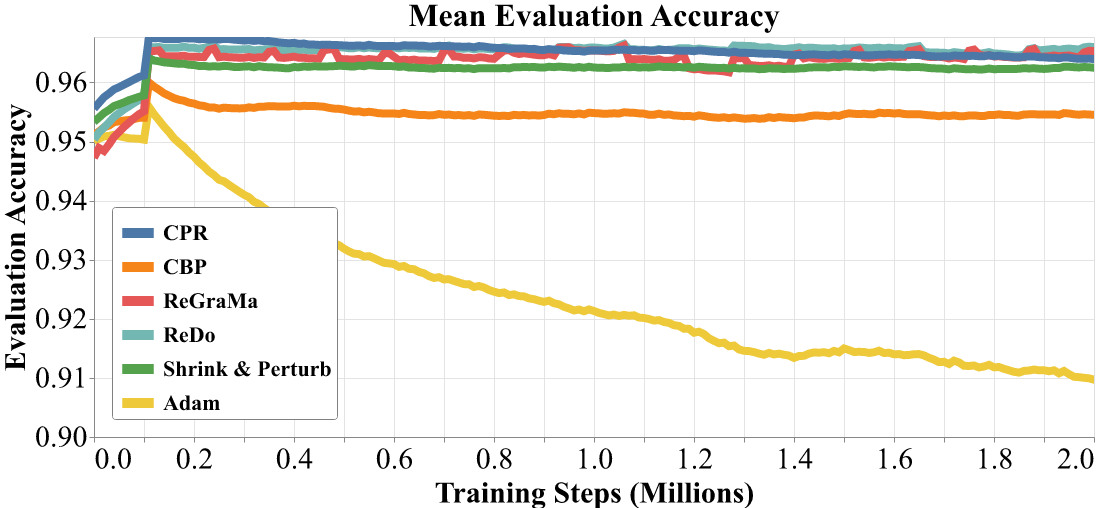}
	\caption{Permuted MNIST Continual Supervised Learning. The Interquartile Mean (IQM) of evaluation accuracy shows CPR maintaining high plasticity and adaptation efficiency throughout the 2 million step training horizon, preventing the performance degradation observed in standard Adam.}
	\label{fig:perm_mnist} 
\end{figure}

We evaluate CPR in the Continual Permuted MNIST setting, a supervised learning regime where the input distribution is non-stationary. In this setup, 200 random permutations are applied to the pixels of MNIST characters sequentially throughout the training process.

As shown in Figure \ref{fig:perm_mnist}, CPR demonstrates the ability to maintain plasticity in the supervised learning regime as well. This experiment tests over 200 tasks, more than any other experiment, and shows CPR matching the stability of reset-based baselines while significantly outperforming standard optimizers like Adam, which exhibit characteristic plasticity loss as the task sequence progresses. The similarity in performance in this experiment demonstrates the interplay between the number of tasks and their complexity when investigating plasticity loss.

\section{Rollout Size}
\label{Appendix:rollout}
Increasing rollout size in on-policy RL is known to stabilize training by reducing gradient approximation noise. While we confirm that sufficiently large rollouts can delay policy collapse in baselines like Adam, relying on massive trajectory buffers masks optimization pathologies rather than solving them. Specifically, we find that large rollouts merely dampen the rate of dormant neuron accumulation without preventing the underlying plasticity loss.

We intentionally adopt a smaller on-policy rollout size to rigorously model the memory and latency constraints of large-scale robotics and high-throughput RL. In these regimes, allocating VRAM to massive buffers is often impractical and competes with the need for high-frequency updates and sample efficiency. Consequently, algorithms that rely on large batches to maintain stability are brittle in resource-constrained environments. Our setup serves as a critical stress test: a robust continual learning algorithm must maintain plasticity even when high-variance gradient estimates are unavoidable.

\begin{figure}[htbp!]
	\centering
	\includegraphics[width=0.8\linewidth]{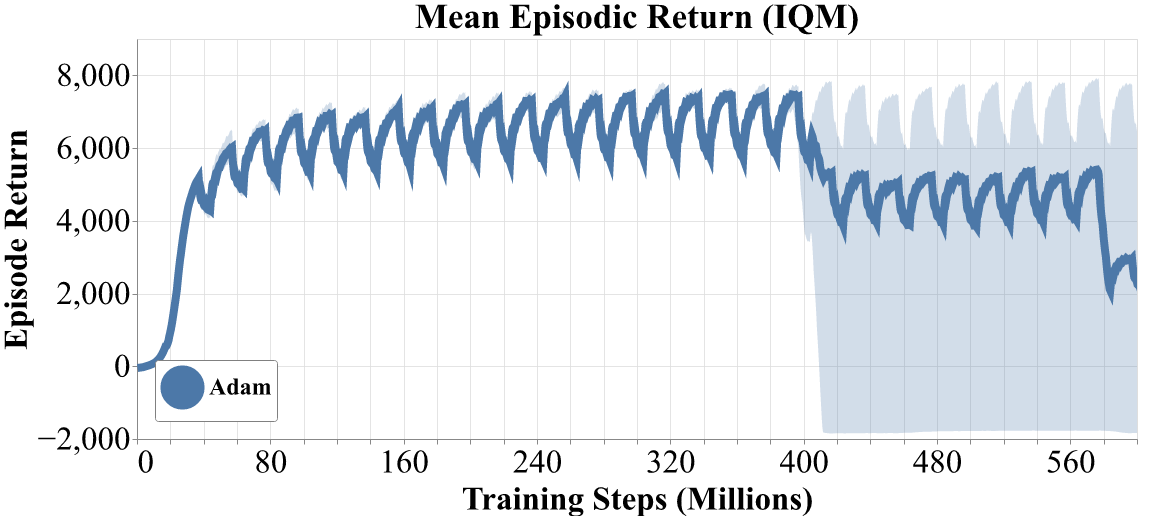}
	\caption{IQM of Adam using larger rollout, with IQR for shaded regions over 15 seeds. This demonstrates large rollouts can delay policy collapse in continual reinforcement learning training}
	\label{fig:bigrollout}
\end{figure}

  \begin{figure}[htbp!]
      \centering
      \begin{subfigure}[t]{0.49\linewidth}
          \centering
	\includegraphics[width=\linewidth]{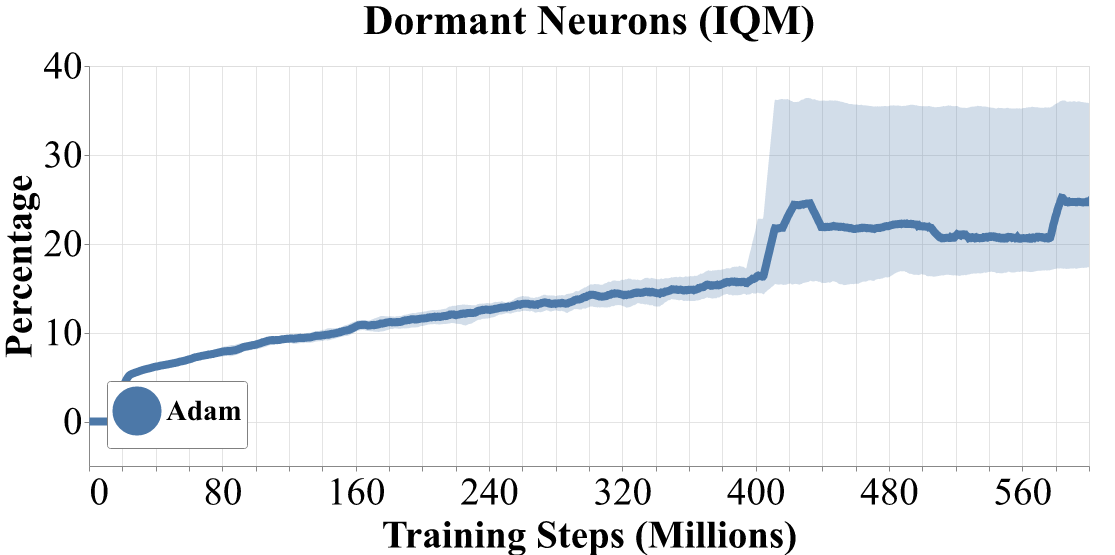}
          \caption{Dormant neurons for Adam with larger rollouts, with IQR for
   shaded regions over 15 seeds. Larger batch sizes delay dormant neuron
  accumulation.}
          \label{fig:big_dormant}
      \end{subfigure}
      \hfill
      \begin{subfigure}[t]{0.49\linewidth}
          \centering
	\includegraphics[width=\linewidth]{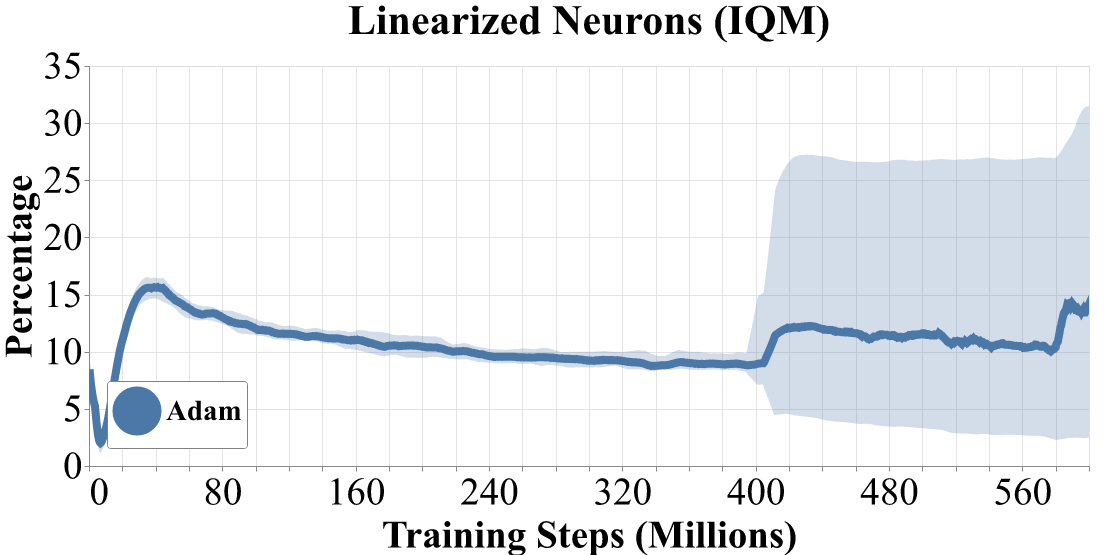}
          \caption{IQM of linearised neurons using Adam and a larger rollout,
  with IQR for shaded regions over 15 seeds. Larger batch sizes delay
  linearised neuron accumulation.}
          \label{fig:big_linear}
      \end{subfigure}
      \caption{Effect of larger rollouts on dormant and linearised neuron
  accumulation.}
      \label{fig:big_batch_neurons}
  \end{figure}


Large rollout sizes reduce the variance of gradient estimation. We demonstrate this empirically in \autoref{fig:bigrollout} where we show how increasing the rollout size can mitigate the policy collapse problem in the SlipperyAnt environment. This motivates the use of dormant reset methods to directly address neuron dormancy under data non-stationarity.

\section{Hyperparameter Configuration}
\label{Appendix:hyperparameter}
\subsection{Continual Ant Experiment Hyperparameter Settings}
The following table summarizes the hyperparameter search space and the optimal values found for each algorithm used in this project. Max reset fraction indicates a maximum proportion of units to be reset as is used by \citet{redo}, where "None" indicates an uncapped reset proportion. Note, different seeds were used for sweeping hyperparameters than generating main results.

\begin{table}[h!]
	\centering
	\caption{PPO Hyperparameters}
	\label{tab:ppo_hyperparameters}
	\begin{tabular}{lc}
		\toprule
		\textbf{Parameter}               & \textbf{Value}     \\
		\midrule
		Learning Rate (SlipperyAnt)      & $10^{-3}$          \\
		Learning Rate (SlipperyHumanoid) & $3 \times 10^{-4}$ \\
		Rollout Steps (SlipperyAnt)      & 196,608            \\
		Rollout Steps (SlipperyHumanoid) & 327,680            \\
		Number of Epochs                 & 4                  \\
		Gradient Steps                   & 32                 \\
		Discount Factor ($\gamma$)       & 0.97               \\
		GAE Lambda ($\lambda$)           & 0.95               \\
		Entropy Coefficient              & $10^{-3}$          \\
		Clip Epsilon                     & 0.2                \\
		Value Function Coefficient       & 0.5                \\
		Normalize Advantages             & True               \\
		Number of Environments           & 2048               \\
		Number of Tasks                  & 20                 \\
		Episode Length                   & 1000               \\
		Steps per Task                   & 20,000,000         \\
		\bottomrule
	\end{tabular}
\end{table}

\begin{table}[h!]
	\centering
	\caption{PPO Network Architecture}
	\label{tab:network_architecture}
	\begin{tabular}{lcc}
		\toprule
		\textbf{Parameter}             & \textbf{Policy Network} & \textbf{Value Function} \\
		\midrule
		Number of Layers               & 4                       & 5                       \\
		Hidden Size (SlipperyAnt)      & 32                      & 256                     \\
		Hidden Size (SlipperyHumanoid) & 128                     & 256                     \\
		Output Size                    & 8                       & 1                       \\
		Activation Function            & Swish                   & Swish                   \\
		Kernel Initialization          & LeCun Normal            & LeCun Normal            \\
		Data Type                      & float32                 & float32                 \\
		\bottomrule
	\end{tabular}
\end{table}

\begin{table*}
	\centering
	\caption{Hyperparameters swept for 5 seeds per configuration. Rankings are based on average final
		performance. Optimal values are reported as SlipperyAnt / SlipperyHumanoid and correspond to the
		values used for the main-paper training results.}
	\label{tab:hyper_sweep}
	\begin{tabular}{llp{5.0cm}p{2.7cm}}
		\toprule
		\textbf{Algorithm} & \textbf{Hyperparameter} & \textbf{Search Space}                                          & \textbf{Selected Values}                              \\
		\midrule
		CPR                & Decay Rate              & [0.9, 0.99]                                                    & 0.99                           /    0.99             \\
		                   & Max Per-unit Reset      & [0.01 to 0.05 (step 0.005; $N=9$)]                             & 0.015                            /  0.05             \\
		                   & Update Frequency        & [100, 1000, 10000]                                             & 1000                     /          1000             \\
		\midrule
		CBP                & Decay Rate              & [0.9, 0.99]                                                    & 0.99         /                      0.99             \\
		                   & Replacement Rate        & \parbox{5cm}{[$10^{-4}, 5\times10^{-4}, 10^{-3},                                                                      \\2.5\times10^{-3}, 3\times10^{-3}, 4\times10^{-3}$]} & $3\times10^{-3}$ / $2.5\times10^{-3}$ \\
		                   & Maturity Threshold      & [100, 1000, 10000]                                             & 100                             /   100              \\
		\midrule
		ReDo               & Score Threshold         & [0.05 to 0.75 (step 0.1; $N$=7)]                               & 0.65                            /   0.5              \\
		                   & Max Reset Fraction      & [None, 0.02, 0.05]                                             & None                          /     None             \\
		                   & Update Frequency        & [100, 1000, 10000]                                             & 100                            /    100              \\
		\midrule
		ReGraMa            & Score Threshold         & [0.05 to 0.75 (step 0.1; $N$=7)]                               & 0.25                              / 0.15             \\
		                   & Max Reset Fraction      & [None, 0.02, 0.05]                                             & None                       /        None             \\
		                   & Update Frequency        & [100, 1000, 10000]                                             & 100                          /      100              \\
		\midrule
		Shrink \&          & Shrink                  & [$10^{-5}$, $10^{-4}$, $10^{-3}$, $5\times10^{-3}$]            & $10^{-3}$                         / $10^{-3}$        \\
		Perturb            & Perturb                 & [$10^{-5}$, $10^{-4}$, $10^{-3}$, $5\times10^{-3}$, $10^{-2}$] & $5\times10^{-3}$                  / $5\times10^{-3}$ \\
		                   & Interval                & [100, 1000, 10000]                                             & 1000                              / 1000             \\
		\bottomrule
	\end{tabular}
\end{table*}

\begin{table}[h]
	\centering
	\caption{Default SAC hyperparameters for Continual MetaWorld experiments}
	\label{tab:cmw_sac_params}
	\begin{tabular}{ll}
		\toprule
		\textbf{Parameter}     & \textbf{Value}       \\
		\midrule
		Learning starts        & 5,000                \\
		Replay ratio           & 8                    \\
		Batch size             & 256                  \\
		Buffer size            & 500,000              \\
		\midrule
		Hidden layers          & 3                    \\
		Hidden size            & 256                  \\
		Activation             & ReLU                 \\
		Dtype                  & float32              \\
		\midrule
		Discount ($\gamma$)    & 0.99                 \\
		Soft update ($\tau$)   & 0.005                \\
		Initial $\alpha$       & 0.2                  \\
		$\alpha$ learning rate & $3 \times 10^{-4}$   \\
		Target entropy         & $-d_{\text{action}}$ \\
		\midrule
		$\log \sigma$ bounds   & $[-20, 2]$           \\
		\midrule
		Num envs               & 10                   \\
		\bottomrule
	\end{tabular}
\end{table}

\begin{table*}[h]
	\centering
	\caption{Hyperparameter sweep ranges and selected values for MetaWorld MT1 experiments, tuned over 5 seeds on Peg Insert Side}
	\label{tab:mt1_reset_params}
	\begin{tabular}{llll}
		\toprule
		\textbf{Method}            & \textbf{Parameter} & \textbf{Search Space}                        & \textbf{Selected Value} \\
		\midrule
		\textbf{CPR}               & Decay Rate         & $[0.9, 0.99]$                                & 0.99                    \\
		                           & Max Per-unit Reset & \parbox{5.5cm}{$[0.01, 0.015, 0.02, 0.025]$} & $0.01$                  \\
		                           & Update Frequency   & $[1000, 5000, 10{,}000]$                     & 1,000                   \\
		\midrule
		\textbf{CBP}               & Replacement Rate   & $[10^{-5}, 10^{-4}, 10^{-3}]$                & $10^{-5}$               \\
		                           & Decay Rate         & $[0.9, 0.99, 0.999]$                         & 0.999                   \\
		                           & Maturity Threshold & $[1000, 5000, 10{,}000]$                     & 1,000                   \\
		\midrule
		\textbf{ReGraMa / ReDo}    & Update Frequency   & $[10^4, 10^5]$                               & $10^5$                  \\
		                           & Score Threshold    & $[10^{-4}, 10^{-3}, 10^{-2}]$                & $10^{-4}$               \\
		                           & Max Reset Fraction & $[None, 0.02, 0.05]$                         & 0.02                    \\
		\midrule
		\textbf{Shrink \& Perturb} & Shrink             & $[10^{-2}, 10^{-3}, 10^{-4}]$                & $10^{-4}$               \\
		                           & Perturbation Std   & $[10^{-3}, 10^{-2}, 10^{-1}]$                & $10^{-3}$               \\
		                           & Interval           & $[1000, 5000, 10{,}000]$                     & 1,000                   \\
		\bottomrule
	\end{tabular}
\end{table*}

\subsection{Continual MinAtar Experiment Hyperparameter Settings}
The Continual MinAtar experiments use discrete SAC with a CNN actor/critic and Adam as the base optimizer. The base Adam optimizer used a learning rate of $3 \times 10^{-4}$ for each baseline. \autoref{tab:minatar_reset_params} reports the selected hyperparameters for each reset method, which were attached on top of Adam via the Optax chain described in \autoref{Appendix:impl}. The SAC optimization settings are the same as \autoref{tab:cmw_sac_params} except with a buffer size of 1M, replay ratio of 4, and 12 parallel envs. Additionally, Continual MinAtar uses a convolutional input layer with 3x3 kernels and stride 1, followed by a single 128-unit hidden layer before the output head.

\begin{table*}[h]
	\centering
	\caption{Selected reset-method hyperparameters for Continual MinAtar experiments.}
	\label{tab:minatar_reset_params}
	\begin{tabular}{llll}
		\toprule
		\textbf{Method}            & \textbf{Parameter} & \textbf{Search Space}                           & \textbf{Selected Value} \\
		\midrule
		\textbf{CPR}               & Decay Rate         & $[0.9, 0.99]$                                   & 0.99                    \\
		                           & Max Per-unit Reset & \parbox{5.5cm}{$[0.01, 0.015, 0.02, 0.025]$}    & 0.015                   \\
		                           & Update Frequency   & $[10^3, 10^4, 10^5]$                            & $10^3$                  \\
		\midrule
		\textbf{CBP}               & Replacement Rate   & $[10^{-6}, 10^{-5}, 10^{-4}]$                   & $10^{-6}$               \\
		                           & Decay Rate         & $[0.9, 0.99, 0.999]$                            & 0.99                    \\
		                           & Maturity Threshold & $[10^3, 10^4, 10^5]$                            & $10^3$                  \\
		\midrule
		\textbf{ReGraMa}           & Update Frequency   & $[10^3, 10^4, 10^5]$                            & $10^{5}$                \\
		                           & Score Threshold    & $[5 \times 10^{-5}, 10^{-4}, 5 \times 10^{-4}]$ & $10^{-4}$               \\
		                           & Max Reset Fraction & $[None, 0.02, 0.05]$                            & 0.02                    \\
		\midrule
		\textbf{ReDo}              & Update Frequency   & $[10^3, 10^4, 10^5]$                            & $10^{5}$                \\
		                           & Score Threshold    & $[5 \times 10^{-5}, 10^{-4}, 5 \times 10^{-4}]$ & $10^{-4}$               \\
		                           & Max Reset Fraction & $[None, 0.02, 0.05]$                            & 0.02                    \\
		\midrule
		\textbf{Shrink \& Perturb} & Shrink             & $[10^{-3}, 10^{-4}, 10^{-5}]$                   & $10^{-5}$               \\
		                           & Perturbation Std   & $[10^{-3}, 10^{-4}, 10^{-5}]$                   & $10^{-4}$               \\
		                           & Interval           & $[10^{3}, 10^{4}, 10^{5}]$                      & $10^{3}$                \\
		\bottomrule
	\end{tabular}
\end{table*}

\subsection{Continual MetaWorld Experiment Hyperparameter Settings}
The Continual MetaWorld experiments use SAC with Muon as the base optimizer. Muon used a learning rate of $10^{-4}$ for each baseline. \autoref{tab:cmw_reset_params} reports the selected hyperparameters for each reset method. The SAC hyperparameters are provided in \autoref{tab:cmw_sac_params}.

\begin{table*}[h]
	\centering
	\caption{Hyperparameter sweep ranges and selected values for Continual MetaWorld experiments, tuned over 5 seeds}
	\label{tab:cmw_reset_params}
	\begin{tabular}{llll}
		\toprule
		\textbf{Method}            & \textbf{Parameter} & \textbf{Search Space}                        & \textbf{Selected Value} \\
		\midrule
		\textbf{CPR}               & Decay Rate         & $[0.9, 0.99]$                                & 0.99                    \\
		                           & Max Per-unit Reset & \parbox{5.5cm}{$[0.01, 0.015, 0.02, 0.025]$} & 0.015                   \\
		                           & Update Frequency   & $[10^3, 10^4, 10^5]$                         & $10^3$                  \\
		\midrule
		\textbf{CBP}               & Replacement Rate   & $[10^{-6}, 10^{-5}, 10^{-4}]$                & $10^{-6}$               \\
		                           & Decay Rate         & $[0.9, 0.99, 0.999]$                         & 0.99                    \\
		                           & Maturity Threshold & $[10^3, 10^4, 10^5]$                         & $10^4$                  \\
		\midrule
		\textbf{ReGraMa}           & Update Frequency   & $[10^3, 10^4, 10^5]$                         & $10^5$                  \\
		                           & Score Threshold    & $[10^{-4}, 10^{-3}, 10^{-2}]$                & $10^{-4}$               \\
		                           & Max Reset Fraction & $[None, 0.02, 0.05]$                         & 0.02                    \\
		\midrule
		\textbf{ReDo}              & Update Frequency   & $[10^3, 10^4, 10^5]$                         & $10^5$                  \\
		                           & Score Threshold    & $[10^{-4}, 10^{-3}, 10^{-2}]$                & $10^{-4}$               \\
		                           & Max Reset Fraction & $[None, 0.02, 0.05]$                         & 0.02                    \\
		\midrule
		\textbf{Shrink \& Perturb} & Shrink             & $[10^{-3}, 10^{-4}, 10^{-5}]$                & $10^{-5}$               \\
		                           & Perturbation Std   & $[10^{-3}, 10^{-4}, 10^{-5}]$                & $10^{-4}$               \\
		                           & Interval           & $[10^{3}, 10^{4}, 10^{5}]$                   & $10^3$                  \\
		\bottomrule
	\end{tabular}
\end{table*}

\section{Continuous Transformation choice}
\label{Appendix:transformations}
In the CPR algorithm, we map the utility score $u_i^\ell$ of a neuron to a partial reset fraction $r_i^\ell$. This mapping is controlled by a transformation function $\phi: \mathbb{R} \to [0, 1]$, a sharpness parameter $\kappa$, a threshold (typically centered at the layer mean, $u=1$), and a maximum per-unit reset fraction $\rho$.

The update rule is defined as:
\begin{equation}
	r_i^\ell = \rho \cdot \phi(u_i^\ell)
\end{equation}

We evaluate four candidate shapes for $\phi(u)$. To ensure fair comparison, all functions are normalized such that $\phi(1) = 1$. That is, a neuron with utility exactly at the threshold receives the maximum reset penalty $\rho$. As utility increases, $\phi(u)$ decays toward 0.

The functions are defined as follows:

\begin{align}
	\label{eq:transformations}
	\phi_{\mathrm{Exp}}(u)      & = \min\!\left(\exp\left[-\kappa (u-1)\right],\,1\right)                                   \\
	\phi_{\mathrm{Sigmoid}}(u)  & = \min\!\left(2 \cdot \sigma\left[-\kappa (u-1)\right],\,1\right)                         \\
	\phi_{\mathrm{Softplus}}(u) & = \min\!\left(\frac{\operatorname{softplus}\left[\kappa (1 - u)\right]}{\ln 2},\,1\right) \\
	\phi_{\mathrm{Linear}}(u)   & = \operatorname{clip}\left(1 - \kappa (u-1),\, 0,\, 1\right)
\end{align}

where $\sigma(z) = (1+e^{-z})^{-1}$ is the logistic sigmoid, $\operatorname{softplus}(z)=\ln(1+e^{z})$, and $\operatorname{clip}(x, a, b) = \max(a, \min(x, b))$.

\textbf{Normalization details:}
\begin{itemize}
	\item \textbf{Sigmoid:} We multiply by 2 because $\sigma(0) = 0.5$. At the mean utility $u=1$, the argument is 0, resulting in $2 \cdot 0.5 = 1$.
	\item \textbf{Softplus:} We divide by $\ln 2$ because $\operatorname{softplus}(0) = \ln(1+e^0) = \ln 2$. This ensures that at $u=1$, the fraction simplifies to 1.
\end{itemize}

\begin{figure}[htbp]
	\centering
		\includegraphics[width=0.9\linewidth, trim=0 0 0 31, clip]{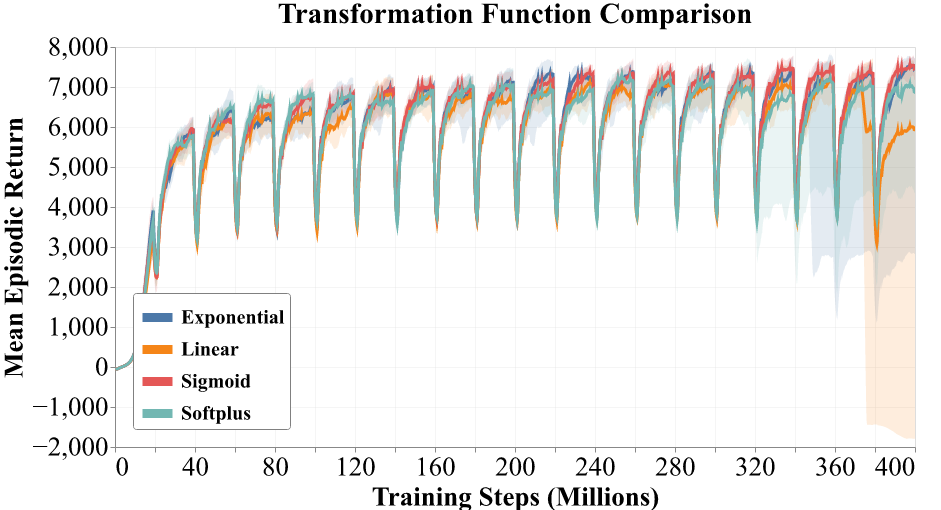}
	\caption{
		Utility transformation function comparison
	}
	\label{fig:util_comp}
\end{figure}

\begin{table}[htbp]
	\centering
	\begin{tabular}{l|cccc}
		\hline
		Transformation & Peak              & Average           & Final             \\
		\hline
		Linear         & 8223.515          & 5778.549          & 5873.256          \\
		Exponential    & 8326.043          & 5999.879          & \textbf{7598.191} \\
		Softplus       & 8145.749          & 5917.624          & 6896.087          \\
		Sigmoid        & \textbf{8360.509} & \textbf{6066.286} & 7498.798          \\
		\hline
	\end{tabular}
	\caption{Training performance on SlipperyAnt (IQM across 15 seeds) with $\rho=0.015$ and $\kappa=16$.}
	\label{tab:transform_perf}
\end{table}

As shown in Table~\ref{tab:transform_perf}, the Sigmoid and Exponential transformations yield the highest returns. To ensure a fair comparison, we calibrated each function to share the same threshold value and fixed $\rho=0.015$.

We observe that the choice of transformation shape has a measurable impact on performance. The Exponential, Sigmoid, and Softplus functions all amplify resets for neurons with \textit{moderately} low gradients, whereas the Linear function is too conservative. The Sigmoid and Exponential functions perform best because their non-linear decay aligns well with the distribution of neuron gradients near the mean; this strengthens the claim for a non-linear transformation.


\begin{figure}[!htbp]
	\centering
	\includegraphics[width=1.0\linewidth, trim=0 0 0 0, clip]{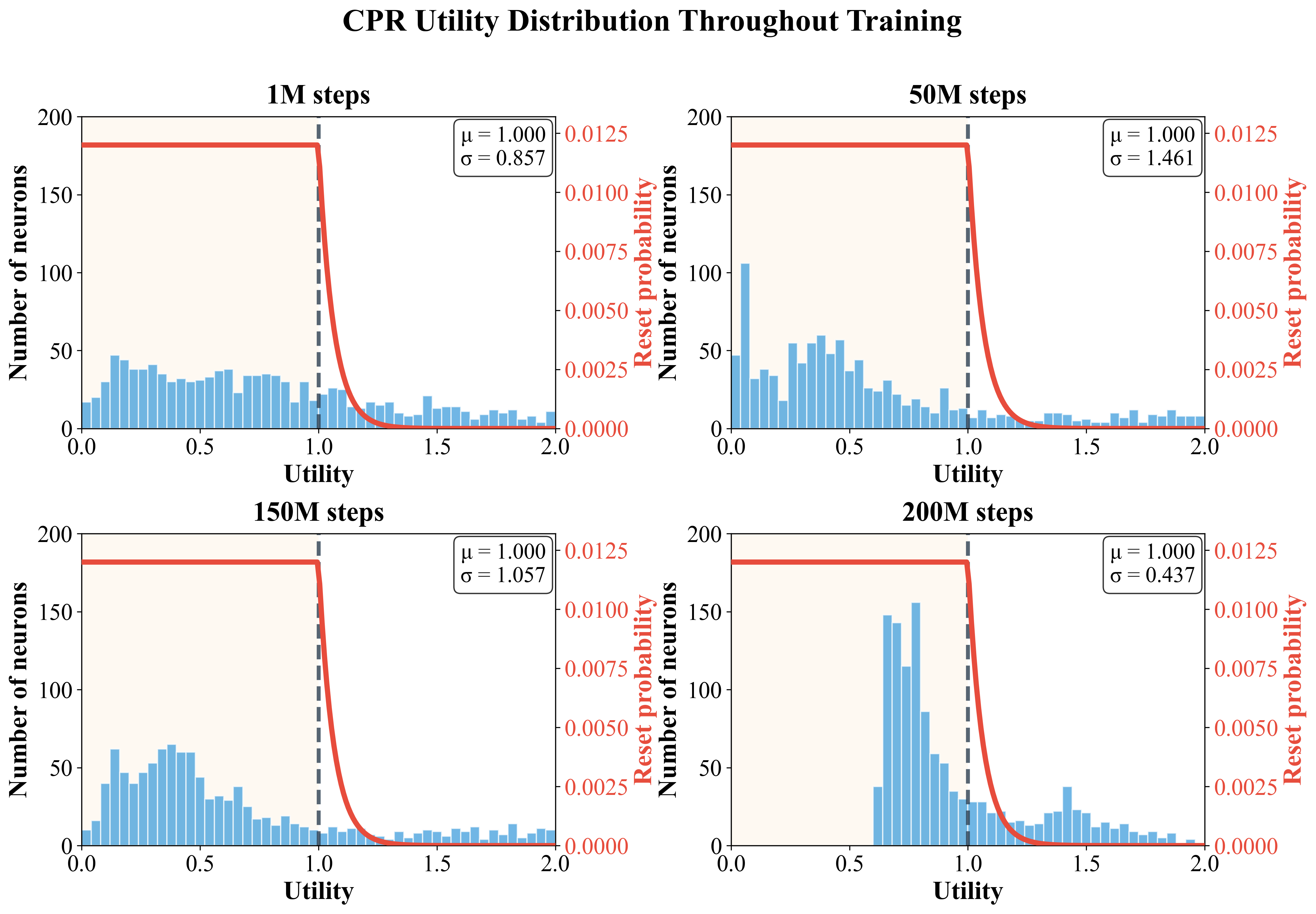}
	\caption{Results from a SlipperyAnt run showing that CPR pushes neurons away from dormant regions}
	\label{fig:util_dist}
\end{figure}

\begin{figure}[!htbp]
	\centering
	\includegraphics[width=0.8\linewidth]{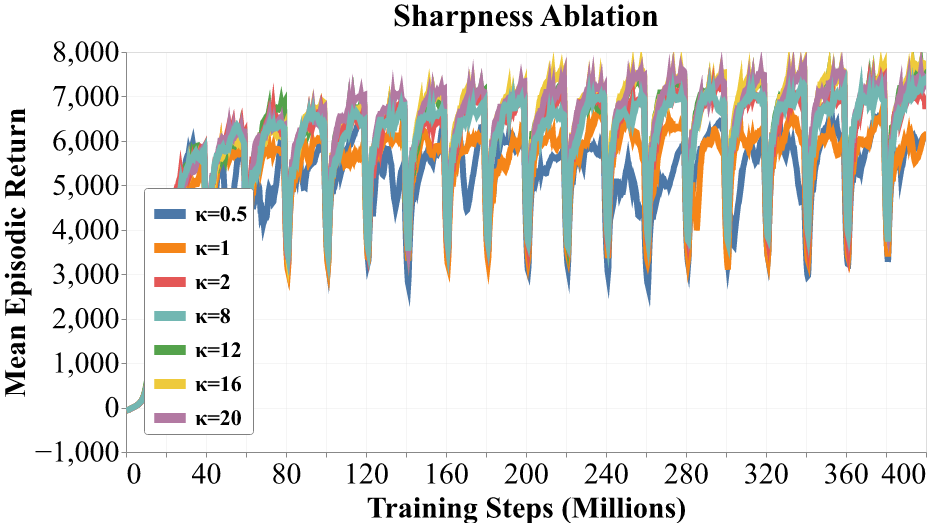}
	\caption{Ablation of the sharpness parameter ($\kappa$) on the SlipperyAnt environment. Performance is highly consistent across a wide range of values ($\kappa$), indicating that CPR is robust to hyperparameter tuning. Notably, low sharpness values ($\kappa \le 1$) lead to significant performance degradation, confirming that resets must be targeted at low-utility neurons rather than applied indiscriminately.}
	\label{fig:sharpness_ablation}
\end{figure}

\section{Implementation Details}
\label{Appendix:impl}
\paragraph{Compatibility with optimizers.}
Other neuron-reset methods additionally reset underlying optimizer state, such as $\mu$ and $\nu$ parameters in Adam \citep{adam}, for the affected low-utility units. We did not find base optimizer parameter resets necessary for our method, which reduces overall implementation complexity.

\paragraph{SlipperyAnt and SlipperyHumanoid.}
For scalable environment parallelism, we build these continual RL environments in Brax \citep{brax}, a JAX \citep{JAX} based physics engine. We implement all baselines and CPR in JAX, which allows compilation of the Continual RL loop end-to-end for faster experimentation, with individual seeds taking approximately an hour on a single Nvidia RTX3090 GPU.

\paragraph{Continual MetaWorld.}
Inspired by ContinualWorld \citep{continualworld} we iterate through MetaWorld \citep{mclean2025meta} tasks sequentially. While ContinualWorld has a focus on catastrophic forgetting we instead measure the adaptation performance using the latest versions of MetaWorld environments. We use the average and final performance of plasticity preservation baselines on this benchmark. This benchmark would take 3 days to run using an RTX3090.

\paragraph{Continual MinAtar.}
The Continual MinAtar benchmark was proposed by \citet{continualminatar} and iterates through three discrete, vision-based Atari tasks (Space Invaders, Asterix and Seaquest). For this benchmark we use discrete SAC with a CNN head for actor and value networks. This benchmark would take 4 hours to run using an RTX3090.


\paragraph{Normalization and gating.}
We form a normalized score $\tilde{u}_i^\ell(t)$ within layer $\ell$ to make selection thresholds comparable across layers and time.

\paragraph{Reset operator.}
When resetting a unit $i$ in layer $\ell$:
\begin{enumerate}[nosep]
	\item \textbf{Incoming weights:} resample the incoming row $W^{\ell}_{i,:}$ (and bias $b^\ell_i$ if present) from the layer’s initial distribution (e.g., Kaiming/He).
	\item \textbf{Outgoing weights:} scale the outgoing column $W^{\ell+1}_{:,i}$ towards \emph{zero}. This prevents immediate disruption to downstream computations and is the standard reset-method design choice, validated in prior work \citep{redo, cbp} .
\end{enumerate}

\paragraph{Reproducibility Statement.}
All of our code, including baselines, environments, experiments, and plotting scripts, is available at: github.com/LucMc/continual-learning/ and open-source under an MIT license. To facilitate reproducibility, the code provides declarative experiment files for generating figures and all project dependencies are specified in the \texttt{pyproject.toml} file at the root of the project, with a \texttt{uv.lock} file pinning them to specific versions used. The code also provides a comprehensive \texttt{README.md} which offers step-by-step instructions for reproducing our results and figures. Finally, we provide full details of hyperparameters used in \autoref{Appendix:hyperparameter}.

\paragraph{Optimizer and Optax chain.}
Our implementation treats reset-methods as optimizer wrappers. We provide an easy-to-use interface for attaching reset methods to optimizer pipelines. Our code is implemented in JAX \citep{JAX} with Optax \citep{Optax}.

Once attached, the Optax optimizer can be used in the same way as a regular Optax optimizer, only if the reset method requires features (i.e. ReDo or CBP) then these are taken as an input to the \texttt{optimizer.update} function.

A simple optimizer setup is shown in \autoref{code:optim}

\begin{minipage}{\linewidth}
	\begin{lstlisting}[label={code:optim}, caption={Demonstration of simple optimizer using a reset method such as CPR}]
from continual_learning.optim import CPR
from continual_learning.utils import attach_reset_method
import optax

# tx can be simple
tx = optax.adam()

# or composite
tx = optax.chain(
    optax.clip_by_global_norm(1.0),
    optax.adam()
)

# ready to be used like any other optax optimizer
tx_w_reset = attach_reset_method(
    ("tx", tx),
    ("reset_method", CPR())
)
\end{lstlisting}
\end{minipage}

\FloatBarrier
\section{Additional Experiments}
\label{Appendix:more_results}

\begin{table*}
	\centering

	\caption{SlipperyAnt average over training, final return (measured at 400M steps for all methods), and peak return. Entries show point estimate with $(+\,/\, -)$ IQR. Best final and best peak are in \textbf{bold}.}
	\label{tab:ant_iqr}

	{\small
		\begin{tabular}{l l l l}
			\toprule
			Method  & IQM Return                    & Final Return (at 400M)        & Peak Return                  \\
			\midrule
			CPR     & \textbf{6051} \((+290/-320)\) & \textbf{7494} \((+230/-235)\) & \textbf{8380} \((+68/-131)\) \\
			CBP     & 4932 \((+576/-873)\)          & 5643 \((+971/-2723)\)         & 7495 \((+426/-787)\)         \\
			ReDo    & 3470 \((+1038/-2434)\)        & 3883 \((+1621/-5731)\)        & 6459 \((+492/-4437)\)        \\
			ReGraMa & 3741 \((+1138/-2717)\)        & 4635 \((+1387/-6458)\)        & 6804 \((+316/-558)\)         \\
			\shortstack{Shrink                                                                                     \\ \& Perturb} & 3469 \((+2733/-2902)\) & -341 \((+7820/-1656)\) & 7812 \((+297/-248)\) \\
			Adam    & 1958 \((+1465/-923)\)         & -1921 \((+92/-77)\)           & 8034 \((+177/-268)\)         \\
			\bottomrule
		\end{tabular}
	}
\end{table*}

\begin{table*}[t]
	\centering

	\caption{SlipperyHumanoid average over training, final return (measured at 400M steps for all methods), and peak return with the step at which it occurs. Entries show point estimate with $(+\,/\, -)$ IQR. Best final and best peak are in \textbf{bold}.
	}
	\label{tab:hum_iqr}

	{\small
		\begin{tabular}{llll}
			\toprule
			Method  & Average Return                & Final Return (at 400M)        & Peak Return                                  \\
			\midrule
			CPR     & \textbf{3270} \((+488/-465)\) & \textbf{4611} \((+425/-443)\) & \textbf{6264} \((+647/-1195)\) at 382M steps \\
			CBP     & 1890 \((+365/-309)\)          & 2487 \((+359/-181)\)          & 2581 \((+962/-797)\) at 222M steps           \\
			ReDo    & 1340 \((+646/-554)\)          & 2021 \((+1189/-1158)\)        & 2439 \((+992/-1203)\) at 322M steps          \\
			ReGraMa & 2393 \((+779/-1180)\)         & 4001 \((+431/-1870)\)         & 4967 \((+1114/-2690)\) at 382M steps         \\
			\shortstack{Shrink                                                                                                     \\ \& Perturb} & 2440 \((+652/-671)\) & 3477 \((+603/-1049)\) & 4268 \((+1123/-1567)\) at 282M steps \\
			Adam    & 1666 \((+994/-838)\)          & 1737 \((+1149/-697)\)         & 3366 \((+2465/-2390)\) at 262M steps         \\
			\bottomrule
		\end{tabular}
	}
\end{table*}

\begin{table}[h]
	\centering
	\caption{Runtime comparison on SlipperyHumanoid. All methods used an identical training setup over 15 seeds on a Nvidia RTX3090. As our implementation captures activations regardless of the reset method, most methods operate at similar computational complexity. The exception being CBP which performs age tracking and reset logic each step.}
	\label{tab:runtime}
	\begin{tabular}{lccc}
		\toprule
		Method            & Median (h) & Overhead \\
		\midrule
		Adam              & 1.17       & ---      \\
		ReGraMa           & 1.22       & +4\%     \\
		Shrink \& Perturb & 1.23       & +5\%     \\
		CPR               & 1.24       & +6\%     \\
		ReDo              & 1.25       & +7\%     \\
		CBP               & 1.34       & +15\%    \\
		\bottomrule
	\end{tabular}
\end{table}

\begin{figure}
	\centering
	\includegraphics[width=0.9\linewidth, trim=0 0 -70 0, clip]{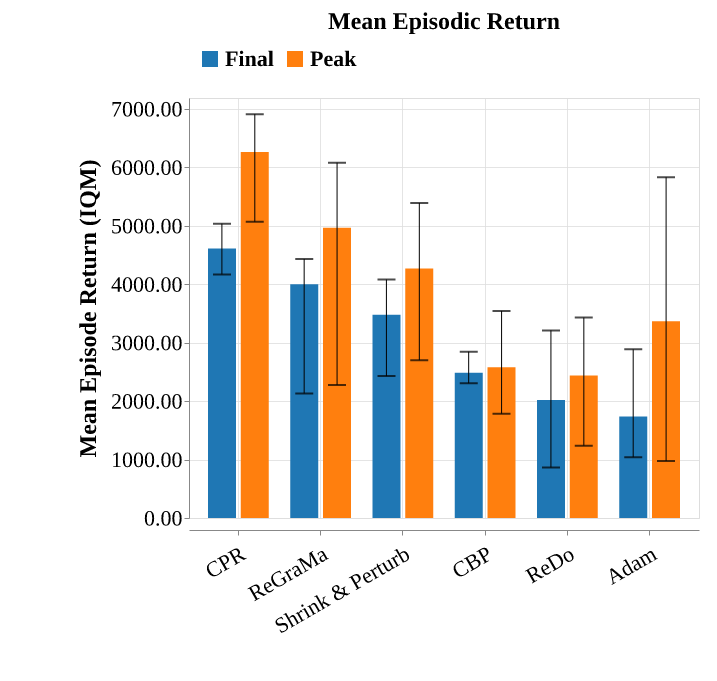}
	\caption{Peak vs.\ final IQM episodic return in SlipperyHumanoid (higher is better).}
	\label{fig:humanoid_peak_vs_final}
\end{figure}

\begin{figure}
	\centering
	\includegraphics[width=\linewidth]{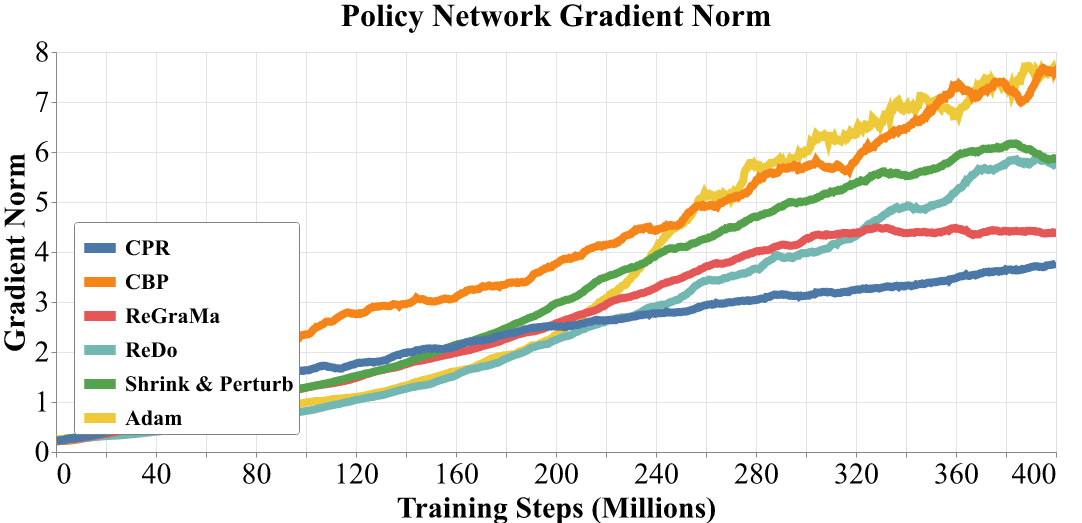}
	\includegraphics[width=\linewidth]{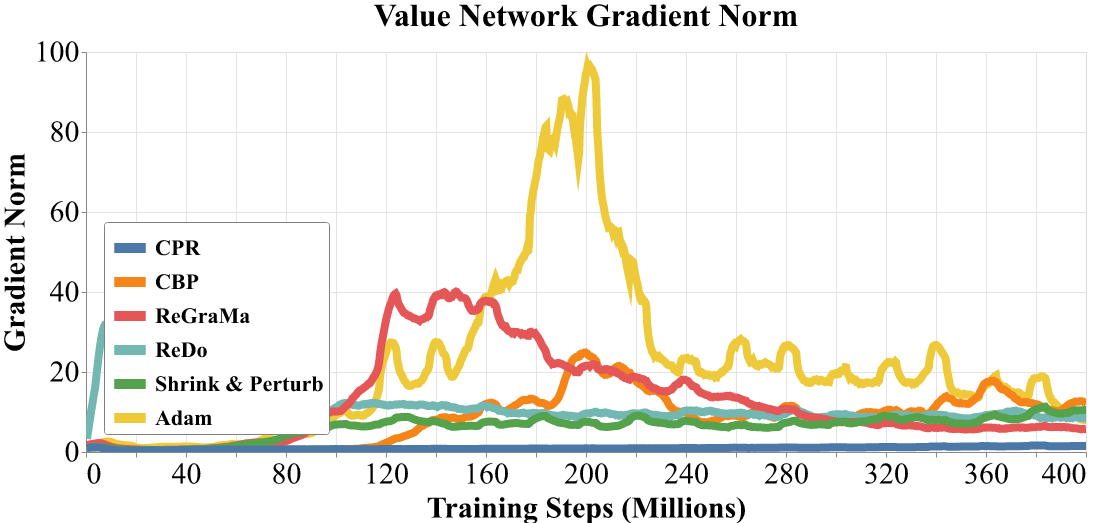}
	\caption{The gradient norm for SlipperyHumanoid, averaged over both actor and critic with IQM across 15 seeds. \textbf{Top}: Policy Network \textbf{Bottom}: Value Network. This demonstrates CPR maintains stable gradient norm throughout training}
	\label{fig:humanoid_grad_norm}
\end{figure}

\begin{figure}
	\centering
	\includegraphics[width=\linewidth]{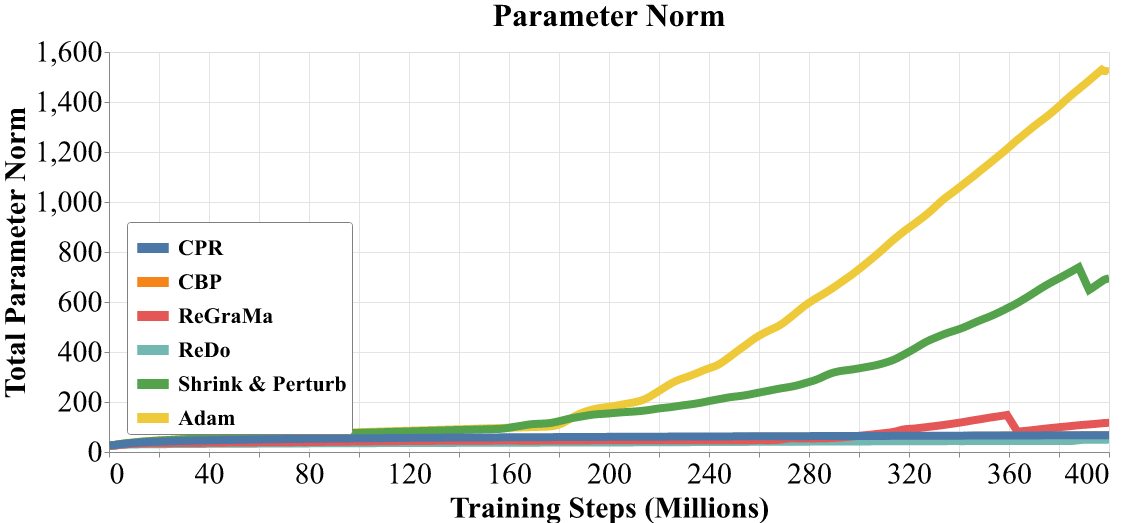}
	\includegraphics[width=\linewidth]{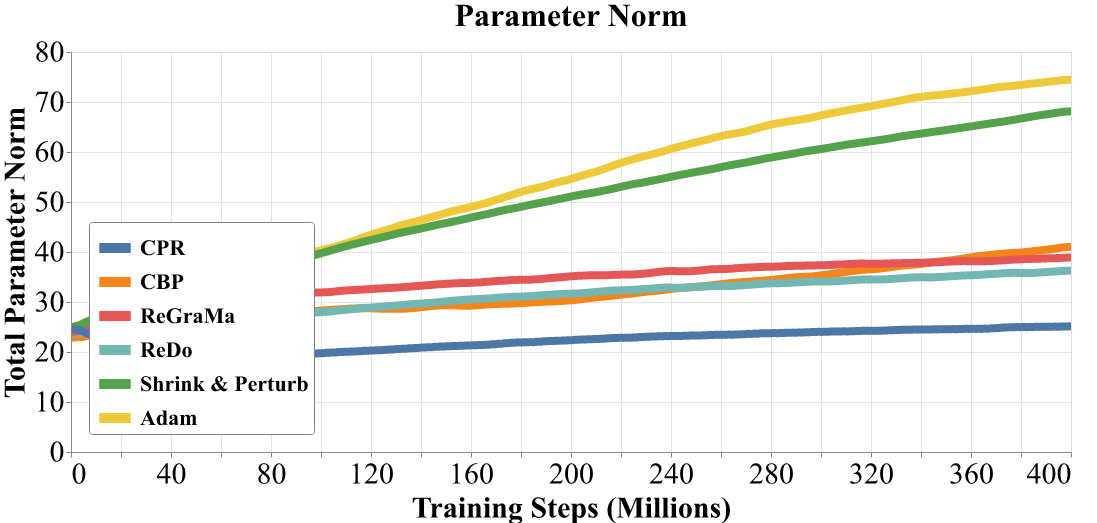}
	\caption{The parameter norm for \textbf{Top}: SlipperyAnt \textbf{Bottom}: SlipperyHumanoid, averaged over both actor and critic with IQM across 15 seeds. This demonstrates that reset methods can prevent parameter-norm growth by periodically refreshing low-utility units.}
	\label{fig:param_norm}
\end{figure}

\begin{figure}
	\centering
	\includegraphics[width=\linewidth]{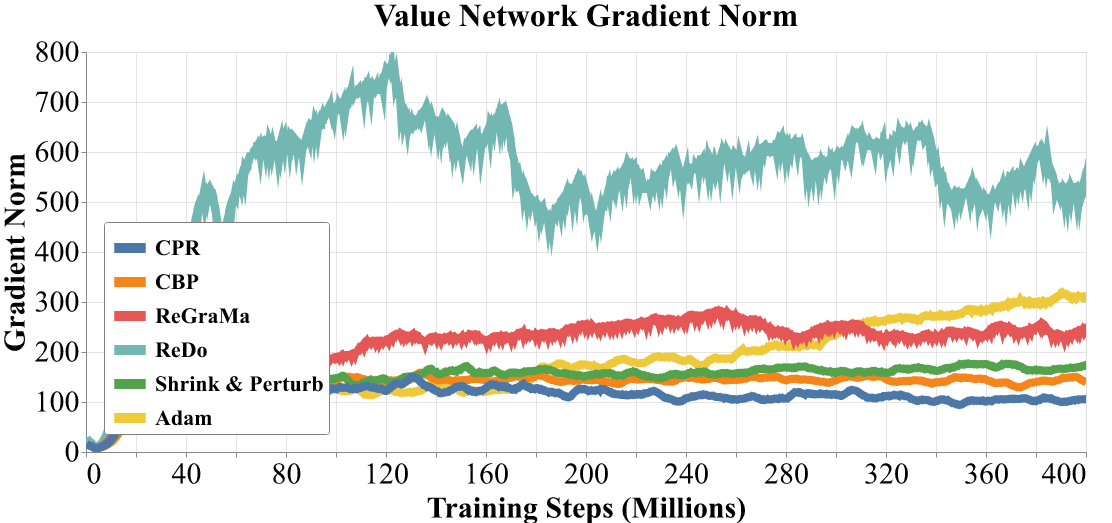}
	\caption{The value network gradient norm for SlipperyAnt, averaged with IQM across 15 seeds.}
	\label{fig:infividual_network_grad_norm}
\end{figure}

\clearpage
\newpage


\end{document}